\documentclass{article}

\PassOptionsToPackage{table,dvipsnames}{xcolor}
\usepackage{iclr2027_conference,times}

\usepackage[utf8]{inputenc}
\usepackage[T1]{fontenc}
\usepackage{hyperref}
\usepackage{url}
\usepackage{booktabs}
\usepackage{amsfonts}
\usepackage{nicefrac}
\usepackage{microtype}
\usepackage{xcolor}
\usepackage{graphicx}
\usepackage{subcaption}
\usepackage{tabularx}
\usepackage{array}
\usepackage{enumitem}
\usepackage{amsmath}
\usepackage{verbatim}
\usepackage{multirow}
\usepackage{multicol}
\usepackage{longtable}
\usepackage{wrapfig}
\usepackage{makecell}

\usepackage{newunicodechar}
\newunicodechar{–}{\textendash}
\newunicodechar{—}{\textemdash}
\newunicodechar{‘}{\textquoteleft}
\newunicodechar{’}{\textquoteright}
\newunicodechar{“}{\textquotedblleft}
\newunicodechar{”}{\textquotedblright}
\newunicodechar{→}{\ensuremath{\rightarrow}}
\newunicodechar{±}{\ensuremath{\pm}}

\hypersetup{
    colorlinks=true,
    linkcolor=red,
    citecolor=teal,
    urlcolor=cyan,
}

\definecolor{MedGemma}{HTML}{F5E8F1}
\definecolor{PanDerm}{HTML}{D8E3EC}
\definecolor{customgreen}{HTML}{008000}

\title{How Medical VLMs Underutilize Their\\Vision Encoders: A Dermatology Perspective}
\author{Janet Wang\textsuperscript{1}\thanks{Corresponding author: \texttt{swang47@tulane.edu}} \quad Yunbei Zhang\textsuperscript{1} \quad Xiao Wang\textsuperscript{2} \quad Jihun Hamm\textsuperscript{1}\\
{\normalfont\small \textsuperscript{1}Tulane University \qquad \textsuperscript{2}Oak Ridge National Laboratory}}

\iclrfinalcopy

\begin{document}

\maketitle
\lhead{Preprint}

\begin{abstract}
  Medical Vision-Language Models (VLMs) show significant promise for clinical image understanding, offering accurate diagnosis with interpretable reasoning. However, a critical performance gap exists between their strong vision encoders and the full multimodal model: in dermatology, the MedSigLIP encoder outperforms MedGemma by an average of 10.26 percentage points even when both use zero target-task labels; few-shot linear probing provides further evidence of strong visual representations. This gap motivates an investigation of how visual information is used in end-to-end diagnosis and why plausible-sounding predictions can lack grounding in image evidence. Using dermatology as our primary testbed, we systematically investigate three hypotheses for this phenomenon. We further provide a mechanistic analysis of the model's internal attention patterns, showing that a simple describe-then-decide prompting strategy increases vision attention by 30–40\% during generation. Task-specific fine-tuning improves dermatology classification but reduces cross-domain medical question-answering performance in our evaluation. To address these challenges, we combine label-free prompting with low-label encoder-assisted reranking while keeping the VLM frozen. We validate the interventions across five VLM backbones in dermatology and provide supporting representation and attention analyses across additional medical modalities.
\end{abstract}

\section{Introduction}
Diagnosing conditions from dermatological images is a challenging task due to inherent complexities, such as subtle variations in disease presentation and the lack of standardized image quality. While advances in computer vision have enabled diagnostic models to achieve expert-level accuracy \citep{Brinkerarticle, Esteva2017DermatologistlevelCO, liu2020deep, soenksen2021using, wang2024achieving}, their deployment in safety-critical domains like healthcare requires more than just precision. It is crucial that a model's decisions are interpretable, ensuring that they are grounded in clinical evidence rather than spurious correlations. Recent emergence of Vision-Language Models (VLMs) in dermatology addresses this need, aiming not only for accurate diagnosis but also for explainable reasoning, thereby paving the way for the reliable, real-world application of medical AI.

The development of dermatological VLMs has often relied on pre-training with large-scale, specialized datasets and advanced architectures. For example, DermLIP is a CLIP-based VLM \citep{radford2021learning} that was pre-trained on Derm1M, a dataset with over one million image-text pairs, enabling tasks like zero-shot classification and concept identification \citep{yan2025derm1m}. Similarly, MONET \citep{kim2024transparent} has fine-tuned a CLIP model on over 100K dermatological images paired with natural language descriptions from a large collection of medical literature, aiming for interpretable diagnoses. Additionally, SkinVL \citep{zeng2025mm} and SkinGPT-4 \citep{zhou2024pre} have integrated large language models into their dermatological VLMs and leveraged large dermatological corpora to facilitate classification with nuanced disease interpretation and visual question answering. More recently, MedGemma, a family of medical foundation models, has emerged to demonstrate superior medical reasoning capabilities \citep{sellergren2025medgemma}. Built upon the powerful Gemma 3 \citep{team2025gemma} architecture and incorporating a medically tuned vision encoder MedSigLIP, MedGemma excels at transparent reasoning, making it an ideal testbed for clinical applications.

\begin{figure*}[t]
    \centering
    \includegraphics[width=1\linewidth]{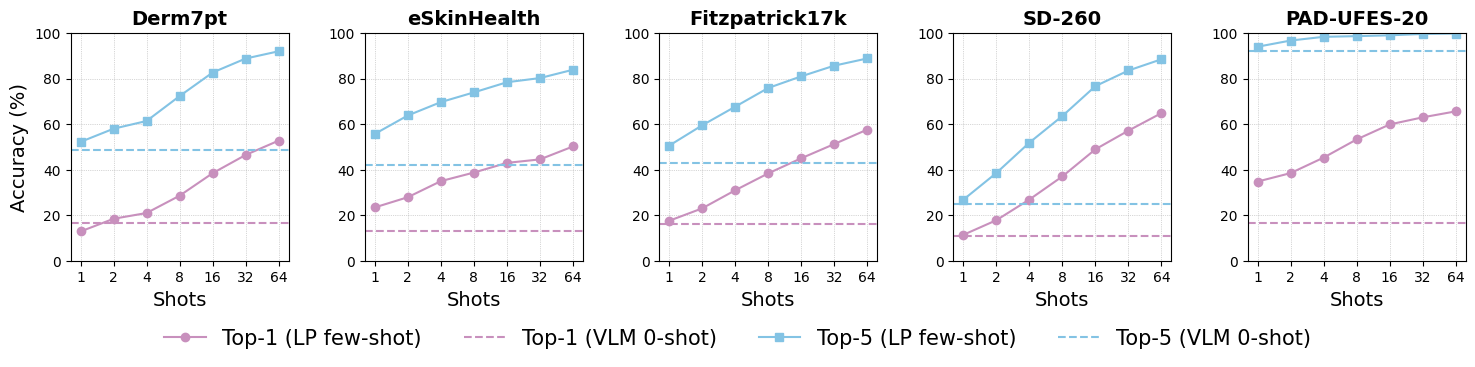}
    \vspace{-4mm}
    \caption{A comparison of the zero-shot performance of the MedGemma-4B VLM against the linear probe performance of its vision encoder, MedSigLIP. LP denotes a linear probe on the vision encoder, while VLM denotes a direct query for diagnosis to the full vision-language model. The comparison shows the diagnostic information accessible through linear probing; the label-free control is reported in Table~\ref{tab:pipeline_performance}.}
    \vspace{-3mm}
    \label{fig:0-shot_MLLM_vs_few-shot_LP}
\end{figure*}

In this work, we identify a critical paradox: while MedGemma’s vision encoder, MedSigLIP, possesses exceptional discriminative power, this strength fails to translate into the zero-shot diagnostic accuracy of the full multimodal model. As shown in Fig.~\ref{fig:0-shot_MLLM_vs_few-shot_LP}, we observe that a simple few-shot linear probe on MedGemma's vision encoder, which is medically tuned from SigLIP \citep{zhai2023sigmoid}, achieves remarkable classification accuracy across five skin disease datasets. Direct comparisons reveal that MedSigLIP is comparable to, and at times superior to, PanDerm \citep{yan2025multimodal}, a specialized dermatology vision encoder, as shown in Appendix Table~\ref{Table_1}. The zero-label comparison in Table~\ref{tab:pipeline_performance} further shows that this gap persists without fitting a classifier on target-task labels. Although fine-tuning improves in-domain accuracy, applying LoRA \citep{hu2022lora} to MedGemma still underperforms a linear probe (Appendix Table~\ref{Table_1}). The Fitzpatrick17k-adapted checkpoint also loses 5.04 and 15.00 percentage points on closed- and open-ended VQA-RAD accuracy, respectively (Appendix~\ref{sec:vqa_transfer}).

This discrepancy leads us to two fundamental questions: \textbf{(1)} What factors contribute to the large performance gap between MedGemma's highly capable vision encoder and its end-to-end zero-shot diagnostic performance? \textbf{(2)} Can we develop inference-time strategies to narrow this gap while keeping the VLM frozen? To answer these questions, our work makes the following contributions:
\begin{enumerate}[leftmargin=*, itemsep=0pt, topsep=0pt]
    \item Using dermatology as our primary testbed, we document a vision--language performance gap under zero target-task labels and compare multimodal image ICL with linear probing and encoder-assisted inference. Additional medical modalities provide supporting evidence of strong frozen visual representations.
    \item We investigate three hypotheses for the gap (\emph{training distribution mismatch}, \emph{vision branch under-reliance}, and \emph{encoder-decoder objective misalignment}), using controlled prompts and attention analysis to relate targeted interventions to changes in model behavior.
    \item Building on these observations, we combine clinical descriptions, DtD, and low-label candidate reranking without fine-tuning the VLM, with improvements across MedGemma, SkinVL, InternVL3, InternVL3.5, and biomed-Qwen in the evaluated dermatology settings.
\end{enumerate}
Our investigation is grounded in dermatology using MedGemma-4B, with additional backbones and medical modalities testing the scope of the observed behavior. The over-reliance on language priors and modality misalignment are fundamental issues facing the broader field of vision-language research \citep{zhai2023investigating, yang2025look, hu2024visual, tong2024eyes}. Therefore, our analysis provides valuable insights into a generalizable approach to improving the reliability and visual grounding of VLMs in safety-critical applications beyond dermatology or healthcare.

\vspace{-1mm}
\section{Related Work}
\vspace{-1mm}
\textbf{Foundation Medical Vision-Language Models for Dermatology.}
Medical VLMs combine specialized data with multimodal architectures \citep{zhang2024generalist, lin2023pmc}. DermLIP uses Derm1M's one million image--text pairs for zero-shot classification and concept identification \citep{yan2025derm1m}; MONET adapts CLIP on over 100K medical image--caption pairs for concept-based diagnosis \citep{kim2024transparent}. Specialized curricula also improve retinal VLMs \citep{holland2025specialized}. SkinVL \citep{zeng2025mm} integrates an LLM using $\sim$10K image--captions and 27K textbook-derived QA pairs, while SkinGPT-4 \citep{zhou2024pre} aligns Llama-2 \citep{touvron2023llama} on over 52K skin disease images. Our study centers on MedGemma \citep{team2025gemma, sellergren2025medgemma}, whose medically tuned MedSigLIP encoder and Gemma~3 language backbone enable analysis of visual representation and end-to-end diagnosis.

\textbf{Misalignment of VLMs and Over-reliance on Language Priors.}
VLMs can underuse their visual representations and rely on language priors \citep{xing2025re}. During generation, declining attention to images can yield visually ungrounded responses \citep{yang2025look}. \citet{tong2024eyes} traced basic visual shortcomings in models such as GPT-4V to the underlying CLIP encoder. Instruction tuning can further degrade visual perception and classification ability \citep{zhai2023investigating}. Automated structured reporting addresses output consistency and hallucination \citep{delbrouck2025automated}. These findings motivate examining visual grounding throughout the multimodal pipeline.

\textbf{Adaptation with Fixed Foundation Models.}
OT-VP and DPCore learn visual prompts for domain alignment and continual adaptation \citep{zhang2025ot, zhang2025dpcore}. BETA uses a local steering model for black-box test-time adaptation \citep{zhang2026adapting}; AReS primes a local encoder through service-model queries before local reprogramming \citep{zhang2026prime}. Multimodal image ICL also supports pathology classification without fine-tuning \citep{ferber2024context}. Our clinical descriptions and DtD use textual context, while encoder-assisted reranking learns a low-label probe and keeps the VLM frozen.

\section{Evaluation Setup and Baseline Comparisons}
\label{sec:evaluation_setup}

\textbf{Models.} The MedGemma family of multimodal models is available in two variants: a 4B and a 27B parameter version. Both models employ the MedSigLIP image encoder, which was pre-trained on a diverse corpus of de-identified medical data, including chest X-rays, dermatology images, ophthalmology images, and histopathology slides. The primary distinction between the two is that the 27B variant was also pre-trained on an additional medical corpus that includes electronic health records (EHRs). As our research focuses on visual diagnosis from images rather than EHR data, MedGemma-4B is a more suitable choice for this study. We formalize the interaction with the VLM $\mathcal{M}$ as $\mathcal{M}[\mathrm{Prompt}, \mathrm{Image}] \rightarrow \mathrm{Response}\;,$
where the model receives a textual prompt and an image of a skin condition as inputs and generates a response containing a diagnosis and an explanation.

\textbf{Datasets.}\label{sec:datasets} To ensure a comprehensive and robust evaluation, our study considers five distinct datasets selected to cover a wide spectrum of diseases and patient demographics. For broad coverage, we include SD-260 \citep{yang2019self}, which contains clinical images across 260 skin conditions, and Fitzpatrick17k \citep{groh2021evaluating}, which features light-skinned images annotated with Fitzpatrick Skin Types. Following the methodology of \citet{wang2025doctor} to leverage their disease checklists, we use the same subset of Fitzpatrick17k as was used in their work. To assess performance on more specific diseases and diverse populations, we incorporate Derm7pt \citep{Kawahara2018-7pt}, focusing on its clinical photos of skin cancer, and eSkinHealth \citep{wang2025eskinhealth}, a specialized dataset of Neglected Tropical Diseases (NTDs) in West African populations. Finally, to test the model's performance on familiar data, we also include PAD-UFES-20 \citep{pacheco2020pad}, an in-domain dataset of skin lesions that was used during MedGemma's pre-training. For datasets that do not provide an official train-test split, we partitioned the data by case and class with a $0.5$ split ratio, with further details provided in Table \ref{Table_1} and Appendix Table \ref{dataset_details}. Implementation details and hyperparameter tuning results for VLM fine-tuning and linear probing are provided in Appendix~\ref{implementation_details}. Additional backbones are InternVL3-4B~\citep{zhu2025internvl3exploringadvancedtraining}, InternVL3.5-4B~\citep{wang2025internvl35advancingopensourcemultimodal}, and biomed-Qwen2.5-VL-3B~\citep{cheng-etal-2025-domain}.

\vspace{-1mm}

\textbf{Label requirements.} Clinical descriptions and DtD use no labeled target images. The Top-5-to-Top-1 route trains a lightweight linear probe on a small labeled support set to shortlist candidates, then uses the frozen VLM for prediction. This route is low-label and VLM-fine-tuning-free. Multimodal image ICL supplies labeled image--class pairs as demonstrations and is distinct from the clinical-description context used in our prompt-only experiments.

\textbf{Zero-label and image-ICL comparisons.} Table~\ref{tab:pipeline_performance} compares MedGemma with MedSigLIP evaluated directly through image--text similarity using the same class labels. MedSigLIP exceeds the VLM on all four evaluated datasets by 7.72--13.18 percentage points (10.26 on average), showing that the gap persists without target-task supervision. The 8-shot image-ICL baseline improves over zero-shot MedGemma by 2.67--4.80 points, while the 8-shot LP has higher accuracy on all four datasets. Appendix~\ref{sec:additional_baselines} includes BiomedCLIP~\citep{zhang2025biomedclipmultimodalbiomedicalfoundation} and DermLIP~\citep{yan2025derm1m} zero-shot comparisons and further protocol details.

\begin{table}[t]
\centering
\small
\setlength{\tabcolsep}{5pt}
\caption{\textbf{The encoder--VLM gap persists without target-task labels.} Top-1 accuracy (\%). The zero-label columns use no fitted classifier; the 8-shot columns compare image ICL, LP, and LP-assisted Top-5-to-Top-1 prediction. A dash marks an unreported setting. LP and reranking variability and the 1-shot results are retained in Appendix Table~\ref{tab:pipeline_performance_detailed}.}
\label{tab:pipeline_performance}
\begin{tabular}{@{}lrrrrr@{}}
\toprule
& \multicolumn{2}{c}{Zero target labels} & \multicolumn{3}{c}{8-shot} \\
\cmidrule(lr){2-3}\cmidrule(l){4-6}
Dataset & VLM & MedSigLIP & Image ICL & LP & Reranking \\
\midrule
Derm7pt        & 16.46 & 24.18 & 19.44 & 28.66 & 35.56 \\
eSkinHealth    & 13.35 & 26.53 & 18.15 & 38.80 & 42.05 \\
Fitzpatrick17k & 16.73 & 28.37 & 20.72 & 38.43 & 37.82 \\
SD-260         & 10.81 & ---   & ---   & 37.12 & 42.33 \\
PAD-UFES-20    & 46.91 & 55.42 & 49.58 & 53.33 & 80.31 \\
\bottomrule
\end{tabular}
\end{table}

\section{Methods}
\vspace{-1mm}
\subsection{Hypothesis 1: Train-test Distribution Mismatch}
\vspace{-1mm}
We first hypothesize that the performance gap stems from a train-test distribution mismatch. That is, the model may not have learned sufficient representations for certain conditions if they were rare or absent in its training data. This distribution mismatch prevents the model from generalizing to these ``tail'' or out-of-distribution classes during zero-shot inference.

\begin{figure*}[t]
    \centering
    \begin{subfigure}[t]{0.49\linewidth}\centering
        \includegraphics[height=1.9in,viewport=0 0 225 216,clip]{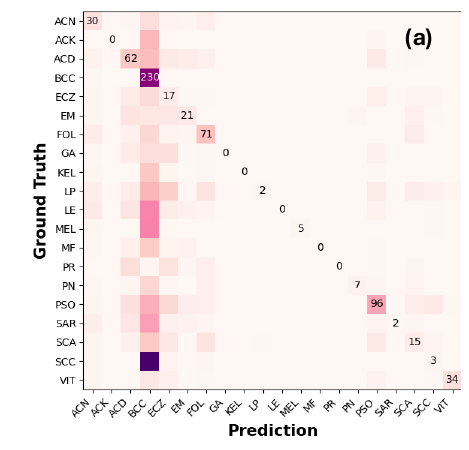}
    \end{subfigure}\hfill
    \begin{subfigure}[t]{0.49\linewidth}\centering
        \includegraphics[height=1.9in,viewport=222 0 447 216,clip]{figures/confusion_matrix.pdf}
    \end{subfigure}
    \vspace{-4mm}
    \caption{\textbf{Clinical context changes the pattern of diagnostic errors.} Fitzpatrick17k confusion matrices for (a) direct answering and (b) clinical descriptions. The full four-strategy comparison, including DtD and 2-shot candidate filtering, appears in Appendix Fig.~\ref{fig:confusion_matrix_full}; class abbreviations are listed in Appendix Table~\ref{tab:fitz_distribution}.}
    \vspace{-3mm}
    \label{fig:confusion_matrix}
\end{figure*}

\textbf{Evidence.} While the exact composition of MedGemma's training data is not public, we can investigate this hypothesis by analyzing the model's prediction bias. The confusion matrix in Fig. \ref{fig:confusion_matrix} clearly demonstrates this bias: the model shows a strong tendency to over-predict certain common conditions, such as basal cell carcinoma (BCC). For instance, nearly all cases of lupus erythematosus (LE) are misclassified as BCC. This raises a critical question: is this error due to a bias in the visual pre-training data (a vision branch issue) or from an incomplete conceptual understanding in the language model (a language branch issue)? By leveraging the model's reasoning capabilities, we can probe its conceptual knowledge. We prompted the model to describe the characteristics of diseases it consistently fails to diagnose (i.e., those with 0\% zero-shot accuracy). As shown in Appendix Table~\ref{tab:bcc_desc}, the model generates detailed descriptions for both BCC and LE, indicating access to relevant disease concepts. These examples motivate the distribution-mismatch hypothesis, while the controlled prompts below test whether class-aligned clinical information improves diagnosis.

\textbf{Solution.} Although the model may lack sufficient visual examples of rare diseases, it possesses a foundational knowledge of general dermatological terms (e.g., ``papules,'' ``inflammation,'' and ``reddish'' ). We can leverage this by re-framing the task from recognizing a disease name to matching visual evidence with a clinical description. Following the methodology of \citet{wang2025doctor, ferber2024context}, our solution provides the model with contextual information in the prompt. Alongside each potential disease name in the list of options, we include a description of its key visual characteristics using a set of generic dermatological vocabularies. Instead of forcing the model to map visual cues to a potentially unfamiliar disease label, this approach encourages it to map visual cues to a provided description. As shown in the improved confusion matrix in Fig. \ref{fig:confusion_matrix} (b), this strategy effectively guides the model toward more accurate and diverse predictions. See visual descriptions in Appendix Table~\ref{tab:condition_descriptions} and complete prompt examples in Appendix Table~\ref{tab:example_prompts}.

\textbf{Context controls.} Class-aligned descriptions outperform both length-matched generic dermatology text and descriptions shuffled across class labels (Table~\ref{tab:context_controls_main}). On Fitzpatrick17k, the three accuracies are 24.01\%, 16.68\%, and 12.71\%; on PAD-UFES-20, they are 52.23\%, 36.36\%, and 15.64\%. Correctly aligned clinical content therefore contributes beyond simply lengthening the prompt.

\begin{table}[t]
\centering
\small
\setlength{\tabcolsep}{8pt}
\caption{\textbf{Clinical content matters beyond prompt length.} Top-1 accuracy (\%) with zero labeled target images. The complete comparison appears in Appendix Table~\ref{tab:context_controls}.}
\label{tab:context_controls_main}
\begin{tabular}{@{}lrr@{}}
\toprule
Prompt context & Fitzpatrick17k & PAD-UFES-20 \\
\midrule
Names only & 16.73 & 46.22 \\
Class-aligned descriptions & 24.01 & 52.23 \\
Length-matched generic text & 16.68 & 36.36 \\
Shuffled descriptions & 12.71 & 15.64 \\
\bottomrule
\end{tabular}
\end{table}



\vspace{-1mm}
\subsection{Hypothesis 2: Under-reliance on the Vision Branch} 
\vspace{-1mm}
\label{Hypothesis 2}
We hypothesize that the VLM's powerful, heavy-weight language model component leads it to over-rely on textual priors when making a diagnosis. This behavior, documented in prior works \citep{yang2025look, tong2024eyes}, causes the model to generate a plausible-sounding diagnosis without sufficiently grounding its prediction and reasoning in the visual evidence of the image.

\begin{figure*}
    \centering
    \includegraphics[width=1\linewidth]{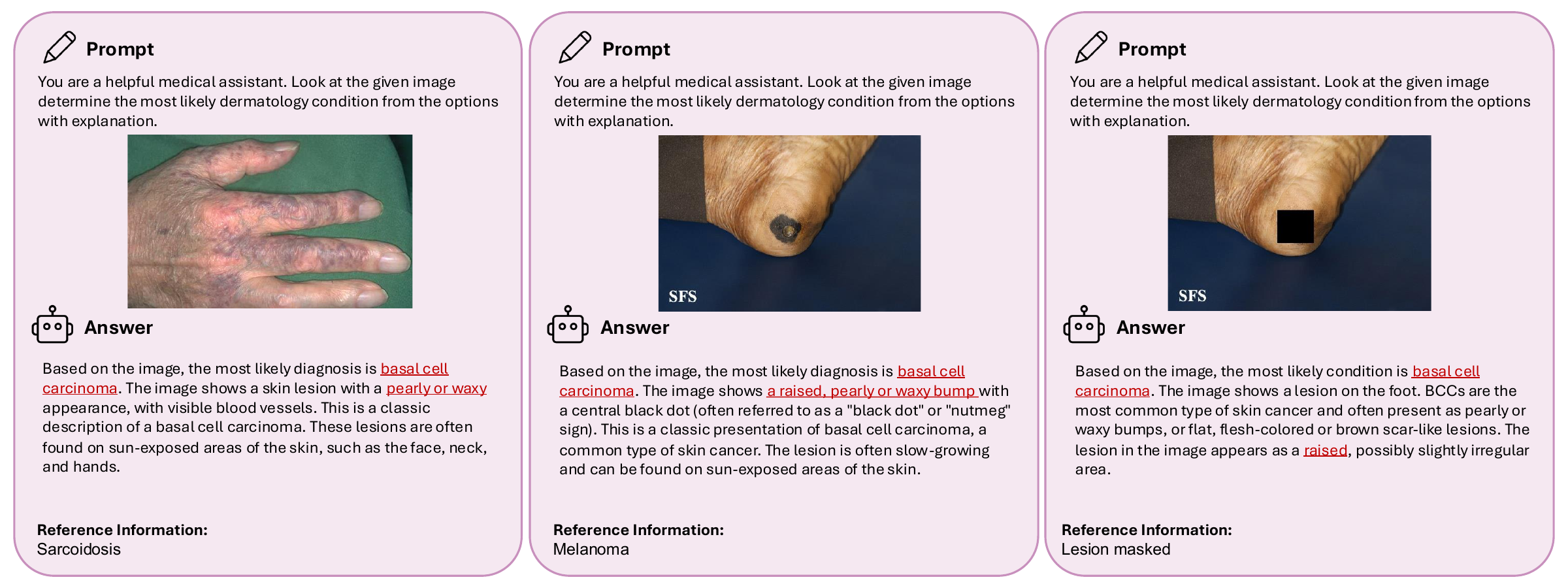}
    \vspace{-4mm}
    \caption{Examples of incorrect answers and hallucinated explanations from MedGemma-4B. The model incorrectly diagnoses sarcoidosis (left) and melanoma (middle) as basal cell carcinoma, providing descriptions of features that are not present in the images. In the third example (right), the lesion is manually masked, yet the model generates the same incorrect diagnosis and explanation, indicating it may generate a prediction first and rationalize it with a post-hoc explanation.}
    \vspace{-3mm}
    \label{fig:hallucination_demo}
\end{figure*}

\textbf{Evidence.} We first observe that the model is prone to hallucinated reasoning, generating explanations that are not grounded in the image. As shown in Fig. \ref{fig:hallucination_demo}, when presented with an image of ``sarcoidosis,'' MedGemma incorrectly identifies the condition as BCC and provides a factually incorrect explanation, as there is no visual evidence of a ``pearly or waxy'' lesion. Similar observations are made for an image of melanoma, even when the lesion is intentionally masked. In these cases, the model tends to provide a generic, post hoc explanation for BCC, indicating that its diagnosis is not derived from the image's visual features. To quantify this behavior, we masked 200 images and tested whether the model would abstain from making a diagnosis when the primary lesion was hidden. As shown in Appendix Fig.~\ref{fig:masked_image_prediction_distribution}(a), the model failed to abstain in 88\% of cases, instead defaulting to a BCC diagnosis. This provides evidence that the model makes predictions without using its ``eyes.''


\textbf{Solution.} To mitigate this over-reliance on language priors, we introduce a ``describe-then-decide'' (DtD) prompting strategy.
This method asks the model to articulate the visible features of the lesion before making a final diagnosis (Appendix Table~\ref{tab:example_prompts}). The premise is that an accurate diagnosis should be preceded by an accurate visual description. This simple intervention yields improved classification accuracy, as shown in the enhanced confusion matrix in Appendix Fig.~\ref{fig:confusion_matrix_full}(c) and Table. \ref{ablation_and_putting_together}. Additionally, when presented again with the 200 masked images, the model's abstention rate increased from 12\% to 72\%, evidencing that this prompt encourages visual grounding.

To examine how DtD changes visual attention, we compare heatmaps for the final diagnosis tokens and for the entire generated response. The examples show more lesion-focused attention under DtD; the heatmaps and their construction are provided in Appendix~\ref{sec:heatmap_construction}. We complement this qualitative comparison with the layer-wise analysis in Sec.~\ref{sec:mechanistic}.

\vspace{-1mm}
\subsection{Hypothesis 3: Misalignment Between the Language and Vision Branches}
\label{sec:description_quality}
\vspace{-1mm}
While DtD improves visual grounding, it does not fully close the performance gap. This leads to our third hypothesis: a fundamental objective misalignment between the VLM's components. The vision encoder is optimized for discriminative tasks (distinguishing classes), whereas the large language model is optimized for generative tasks (explaining and reasoning). We hypothesize that the VLM is better suited for explanation and description than for forced-choice classification and that leveraging these distinct strengths is key to maximizing performance.


\textbf{Evidence.} Previous sections have already established that the vision encoder's discriminative performance surpasses that of the VLM. To complete our evidence, we now evaluate the VLM's primary strength: its ability to generate high-quality, descriptive text. Following the methodology of \citet{wang2025doctor}, we prompted MedGemma-4B to describe lesions using a structured format covering five key clinical criteria: [\textit{Location Site, Lesion Type, Shape/Border, Color, Texture}]. An example of a generated description is shown in Appendix Fig.~\ref{fig:description_quality}. To quantitatively assess quality, the model was instructed to review the generated text against the original image and verify the accuracy of each of the five requested criteria. The resulting self-evaluation score (0--5) counts the criteria that the model judges to be correctly described. As shown in Fig. \ref{fig:single_score_distribution}, the model assigns high scores (predominantly 4 or 5) to its descriptions. This performance stands in stark contrast to the hallucinated and visually ungrounded explanations observed during the standard zero-shot diagnosis (see Fig. \ref{fig:hallucination_demo}), suggesting a difference between descriptive and classification behavior that motivates encoder-assisted candidate selection.


\textbf{Solution.} 
Our solution is a two-stage pipeline that leverages the complementary strengths of the vision and language components. Instead of forcing the VLM to perform a task it is not optimized for, we delegate the initial discriminative work to the vision encoder. First, we use the powerful vision encoder with a few-shot linear probe to identify the top-5 most likely diagnosis candidates for a given image. This narrows the field to a small set of high-probability candidates. Then, we feed these top-5 candidates to the full VLM and prompt it to make the final diagnosis from only that reduced set. We term this strategy as ``Top-5 to Top-1.''

As shown in Table \ref{tab:pipeline_performance}, this approach is highly effective, especially in low-data regimes. With 1 to 8 shots, the two-stage pipeline improves over a linear probe in most evaluated settings; at 8 shots on Fitzpatrick17k, it obtains 37.82\% compared with 38.43\% for LP. The benefit diminishes as the number of shots increases (e.g., 16 or more), as the linear probe becomes powerful enough on its own (see Appendix \ref{ablation_shots}). This makes our approach particularly valuable for real-world medical applications where labeled data is scarce.

This strategy offers three additional benefits. First, by providing the VLM with only 5 options instead of 20 or more, it reduces the prompt length, saving token space for more context. Second, the vision encoder is a plug-and-play component; practitioners can substitute MedSigLIP with any other powerful, domain-specific vision encoder to potentially boost performance further. Third, and perhaps most importantly, we still have the model's reasoning capabilities at our disposal.


\begin{table*}[t]
\centering
\scriptsize
\caption{Ablation study on the effectiveness of each proposed strategy. This table presents Top-1 accuracy for five inference configurations. Clinical descriptions and DtD are label-free; candidate filtering uses target labels. The ``DA'' column serves as the baseline, using a standard classification prompt. The ``in-context'' strategy augments the prompt with clinical descriptions; ``DtD'' requires the model to describe visual features before diagnosis; and ``Top-5 to Top-1'' is our two-stage strategy where the VLM reranks candidates from a 8-shot linear probe. The final ``all combined'' column integrates all proposed strategies.}
\vspace{-1mm}
\label{ablation_and_putting_together}
\tiny
{
\renewcommand{\arraystretch}{0.95}
\resizebox{0.95\textwidth}{!}{%
\begin{tabular}{l|ccccc}
\toprule
\textbf{Dataset} & \textbf{DA} & \textbf{in-context} & \textbf{DtD} & \textbf{top-5 to top-1} & \textbf{all combined} \\
\midrule
Derm7pt          & 16.46 (\tiny $\pm$ 0.17) & 26.72 (\tiny $\pm$ 0.82) & 30.98 (\tiny $\pm$ 1.37) & 35.10 (\tiny $\pm$ 1.95) & 38.27 (\tiny $\pm$ 0.81) \\
eSkinHealth      & 13.35 (\tiny $\pm$ 0.23) & 30.07 (\tiny $\pm$ 1.51) & 29.77 (\tiny $\pm$ 0.92) & 43.94 (\tiny $\pm$ 1.03) & 48.00 (\tiny $\pm$ 0.86) \\
Fitzpatrick17k   & 16.73 (\tiny $\pm$ 0.25) & 24.01 (\tiny $\pm$ 2.01) & 24.75 (\tiny $\pm$ 1.53) & 28.32 (\tiny $\pm$ 1.67) & 40.97 (\tiny $\pm$ 1.41) \\
SD-260 (subset)  & 24.31 (\tiny $\pm$ 0.15) & 34.29 (\tiny $\pm$ 1.06) & 33.80 (\tiny $\pm$ 1.37) & 40.01 (\tiny $\pm$ 2.18) & 44.90 (\tiny $\pm$ 0.91) \\
PAD-UFES-20      & 46.22 (\tiny $\pm$ 0.17) & 52.23 (\tiny $\pm$ 1.26) & 62.33 (\tiny $\pm$ 1.58) & 75.23 (\tiny $\pm$ 2.07) & 84.68 (\tiny $\pm$ 1.80) \\
\bottomrule
\end{tabular}%
}
}
\vspace{-2mm}
\end{table*}

\subsection{Ablation Study and Combined Performance}
Having investigated three hypotheses for the performance gap and proposed a corresponding intervention for each, we now conduct an ablation study to evaluate the individual and combined effects of our strategies. We compare the following five inference configurations, with the results presented in Table. \ref{ablation_and_putting_together}. For the ``Top-5 to Top-1'' and ``all combined'' configurations, we use a 8-shot linear probe to generate the initial candidates. We choose 8 shots because, as shown in Table \ref{table:top5-to-top1}, it represents a point of high efficacy in the low-data regime before performance gains begin to saturate. For the SD-260 dataset, providing expert-verified clinical descriptions for all 260 classes was infeasible. Therefore, for this study, we use a subset of SD-260 containing only the classes that overlap with the other datasets. Example prompts are provided in Appendix Table~\ref{tab:example_prompts}.

The results of the ablation study provides evidence for the effectiveness of the proposed fine-tuning-free pipeline. We observe that each of the three strategies (in-context, DtD, and ``Top-5 to Top-1'') independently yields a significant improvement in Top-1 accuracy over the direct answer baseline across all five datasets. These improvements support the practical utility of the three complementary interventions. Most importantly, the ``all combined'' design consistently achieves the highest performance, substantially outperforming any single strategy. The combined results show that the interventions can work together to narrow the performance gap and enhance diagnostic accuracy.

\begin{table}[t]
    \centering

    \caption{Generalizability on SkinVL (LLaVA-based). Top-1 Accuracy (\%) comparison. Our proposed inference pipeline consistently improves performance over the direct answer baseline, demonstrating robustness across different VLM architectures.}
    \label{tab:skinvl_main}
    \tiny
    \resizebox{0.8\textwidth}{!}{
    \begin{tabular}{l ccc}
        \toprule
        \textbf{Inference Strategy} & \textbf{Derm7pt} & \textbf{eSkinHealth} & \textbf{Fitzpatrick17k} \\
        \midrule
        Direct Answer (Baseline) & 12.77 (\tiny $\pm$ 0.12) & 23.76 (\tiny $\pm$ 0.21) & 10.05 (\tiny $\pm$ 0.08) \\
        + Top-5 Filtering & 29.66 (\tiny $\pm$ 1.45) & 40.25 (\tiny $\pm$ 1.82) & 58.37 (\tiny $\pm$ 1.15) \\
        + In-Context Description & 31.19 (\tiny $\pm$ 1.33) & 44.03 (\tiny $\pm$ 2.01) & 62.97 (\tiny $\pm$ 1.28) \\
        \textbf{+ DtD (Ours)} & \textbf{32.56} (\tiny $\pm$ 0.95) & \textbf{44.89} (\tiny $\pm$ 1.67) & \textbf{63.91} (\tiny $\pm$ 1.05) \\
        \bottomrule
    \end{tabular}
    }
    \vspace{-2mm}
\end{table}

\subsection{Generalization to Other VLMs for Dermatology}
To demonstrate that the identified challenges and proposed solutions are not specific to the MedGemma architecture, we extended our evaluation to SkinVL \citep{zeng2025mm}. As a LLaVA-based model, SkinVL represents a distinct architectural lineage from MedGemma. First, we confirmed that the performance gap is a pervasive issue: consistent with MedGemma, a simple linear probe on SkinVL's vision encoder significantly outperforms its zero-shot VLM accuracy (see Appendix \ref{skinvl} for full comparison). Second, we evaluated our fine-tuning-free pipeline on SkinVL. Due to its stricter context window, we adapted the pipeline to perform ``Top-5 to Top-1'' filtering first. As shown in Table~\ref{tab:skinvl_main}, our strategies yield consistent performance gains across multiple datasets, even improving upon the high baselines of Fitzpatrick17k (a dataset included in SkinVL's pre-training). This supports transfer to the LLaVA-based SkinVL architecture. We further evaluate InternVL3-4B, InternVL3.5-4B, and biomed-Qwen2.5-VL-3B on Derm7pt and Fitzpatrick17k (Table~\ref{tab:additional_backbones}). DtD improves over direct answering in all six model--dataset pairs, and candidate-assisted inference gives the highest accuracy among the three settings. On InternVL3.5, Top-5-to-Top-1 improves over DA by 11.11 and 12.91 percentage points, respectively. Appendix~\ref{sec:additional_models} provides the extended comparison.

\begin{table}[t]
\centering
\small
\setlength{\tabcolsep}{6pt}
\caption{\textbf{The interventions transfer to additional VLM backbones.} Top-1 accuracy (\%) on Derm7pt (D7) and Fitzpatrick17k (F17). DA: direct answer; DtD: describe-then-decide; Top-5: 8-shot candidate-assisted Top-5-to-Top-1 inference. Full results are listed in Appendix Table~\ref{tab:additional_backbones_full}.}
\label{tab:additional_backbones}
\begin{tabular}{@{}llrrr@{}}
\toprule
Model & Dataset & DA & DtD & Top-5 \\
\midrule
InternVL3-4B & D7 & 5.68 & 6.20 & 14.83 \\
             & F17 & 10.10 & 12.23 & 19.70 \\
InternVL3.5-4B & D7 & 7.21 & 7.85 & 18.32 \\
               & F17 & 9.25 & 13.15 & 22.16 \\
biomed-Qwen2.5-VL-3B & D7 & 20.93 & 38.50 & 41.24 \\
                     & F17 & 12.14 & 19.96 & 22.35 \\
\bottomrule
\end{tabular}
\end{table}

\vspace{-1mm}
\section{Mechanistic Analysis of Vision Attention}
\vspace{-1mm}
\label{sec:mechanistic}
Inference-time strategies narrow the vision--language gap, but they do not explain why the gap exists in the first place. Why does the model underutilize its vision encoder? To answer this, we analyze internal attention patterns during autoregressive generation.

\phantomsection
\label{sec:var}
\textbf{Vision Attention Ratio.}
We define the Vision Attention Ratio (VAR) as the fraction of attention allocated to vision tokens during generation. For decoder layer~$l$ and generation step~$t$:
\[
\text{VAR}_{l,t}
    = \frac{\sum_{i \in \mathcal{V}} A_{l,t,i}}
           {\sum_{j} A_{l,t,j}}\,,
\]
where $A_{l,t}$ is the head-averaged attention distribution of the newly generated token over all key--value positions, and $\mathcal{V}$ indexes vision-token positions. Per-layer VAR is averaged over all generation steps and test images, with higher values indicating more attention allocated to visual tokens. Two prompting conditions are compared: \textbf{Direct Answer (DA)}, where the model classifies directly, and \textbf{Describe-then-Decide (DtD)}, where the model first describes the image and then classifies. Full attention weights are extracted using eager attention across all 34 transformer layers of MedGemma-4B, with 100 test images per dataset.

\begin{figure*}[t]
    \centering

    \begin{subfigure}[t]{0.31\textwidth}
        \centering
        \includegraphics[width=\linewidth]{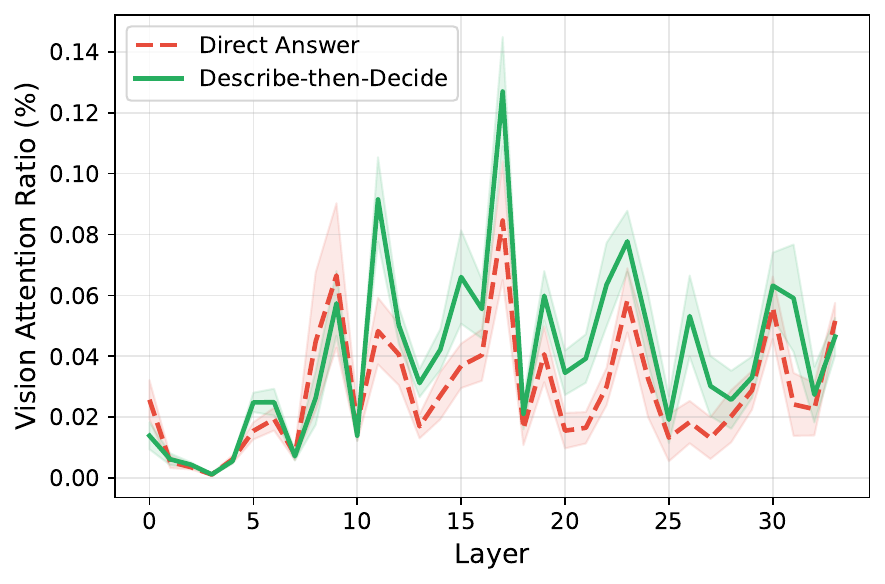}
        \caption{Fitz17k per-layer VAR}
        \label{fig:var_fitz}
    \end{subfigure}
    \hfill
    \begin{subfigure}[t]{0.31\textwidth}
        \centering
        \includegraphics[width=\linewidth]{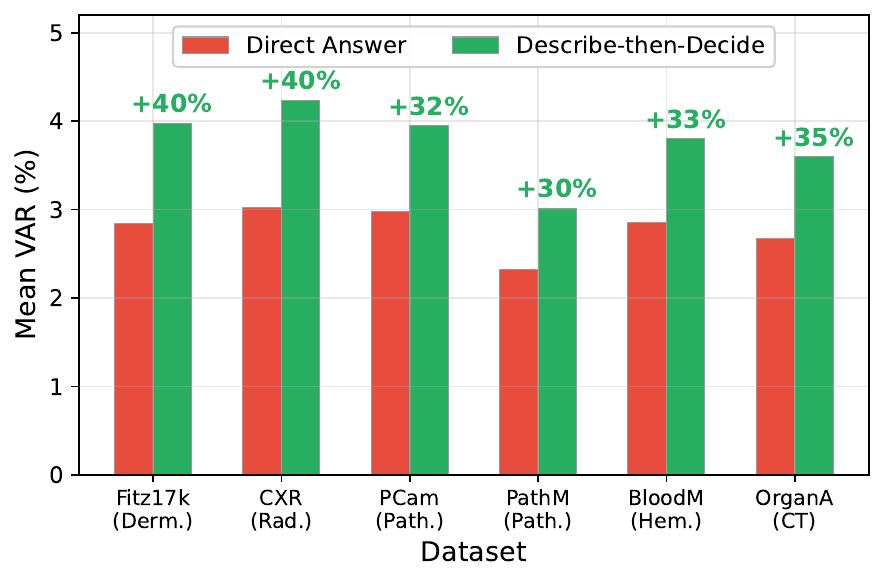}
        \caption{Mean VAR across modalities}
        \label{fig:var_bar_comparison}
    \end{subfigure}
    \hfill
    \begin{subfigure}[t]{0.32\textwidth}
        \centering
        \includegraphics[width=\linewidth]{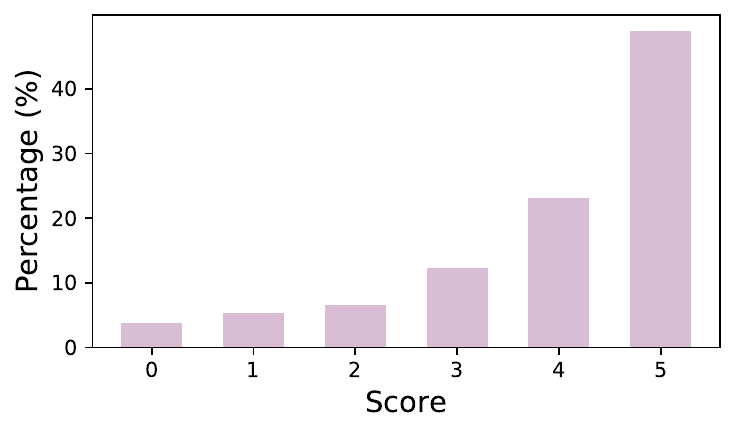}
        \caption{Description quality scores}
        \label{fig:single_score_distribution}
    \end{subfigure}

    \caption{
    Vision Attention Ratio (VAR) and description quality scores.
    \textbf{(a)} Per-layer VAR for Fitz17k shows that Describe-then-Decide (DtD) increases vision attention across all layers, with the most pronounced gains at global attention layers 5, 10, 17, 22, 27, and 33.
    \textbf{(b)} Mean VAR across six medical imaging datasets shows that DtD increases attention allocated to visual tokens by 30--40\%.
    \textbf{(c)} Distribution of 0--5 quality scores for 100 generated descriptions, based on the five clinical criteria described in Sec.~\ref{sec:description_quality}.
    }
    \label{fig:var_and_human_eval}
\end{figure*}

\textbf{DtD Increases Vision Attention Across Modalities.}
On Fitz17k (Fig.~\ref{fig:var_fitz}), DA allocates only 2.85\% of generation-time attention to vision tokens, indicating that the model largely ignores the image when classifying directly. DtD raises this to 3.98\%, a relative gain of 40\%. The effect generalizes: Fig.~\ref{fig:var_bar_comparison} shows DtD producing 30--40\% higher mean VAR than DA across all six modalities, with the largest gains in radiology (CheXpert, +40\%) and dermatology (Fitz17k, +40\%). The mechanism is straightforward. The description phase forces the model to continuously consult vision tokens, whereas under DA a classification label can be produced from language priors alone. The increased visual attention accompanies the accuracy gains reported in Sec.~\ref{Hypothesis 2}.
The per-layer profile (Fig.~\ref{fig:var_fitz}) reveals periodic VAR spikes at layers 5, 10, 17, 22, 27, and 33. These match the global attention layers in Gemma~3's sliding-window architecture~\citep{team2025gemma}; the remaining layers use local sliding-window attention and attend minimally to vision tokens. DtD increases VAR most at these global layers. Attention entropy tells the same story: DtD produces lower entropy than DA across all six modalities, indicating more focused and selective attention (Appendix~\ref{sec:entropy_analysis}).

\phantomsection
\label{sec:cross_modality}
\textbf{Generality Across Medical Modalities.}
We further evaluate five additional medical modalities: CheXpert (CXR)~\citep{irvin2019chexpert}, a 5-class chest X-ray dataset; PatchCamelyon (PCam)~\citep{Veeling2018-qh}, binary histopathology; and three MedMNIST~\citep{medmnistv2} benchmarks, PathMNIST (PathM, 9-class colorectal pathology), BloodMNIST (BloodM, 8-class hematology), and OrganAMNIST (OrganA, 11-class abdominal CT).
\emph{The gap extends across the evaluated modalities.}
Table~\ref{tab:cross_modality} shows that 8-shot linear probing of the frozen encoder outperforms zero-shot MedGemma on all five additional datasets. These comparisons extend the representation-level observation beyond dermatology; the zero-label control is evaluated on the dermatology datasets in Table~\ref{tab:pipeline_performance}.
\emph{DtD increases visual attention across modalities.}
The same layer-wise VAR pattern observed in dermatology holds across all modalities (Appendix Fig.~\ref{fig:var_layers}). Radiology and dermatology see the largest gains (+40\%), while pathology and hematology see +30--35\%. This likely reflects the fact that modalities requiring whole-image spatial reasoning benefit most from forced visual description.




\begin{table}[t]
\centering
\caption{Few-shot linear probe accuracy (\%) on frozen MedSigLIP features
vs.\ MedGemma-4B VLM zero-shot classification across medical imaging
modalities. LP results are averaged over 5 runs. At 8 shots, LP outperforms the zero-shot VLM on every dataset.}
\label{tab:cross_modality}
\tiny
{
\renewcommand{\arraystretch}{0.95}
\resizebox{\linewidth}{!}{%
\begin{tabular}{l l c c c c c | c}
\toprule
 & & & \multicolumn{4}{c|}{\textbf{Linear Probe}} & \textbf{VLM} \\
\cmidrule(lr){4-7} \cmidrule(l){8-8}
\textbf{Dataset} & \textbf{Modality} & \textbf{\#Cls}
  & \textbf{1-shot} & \textbf{4-shot}
  & \textbf{8-shot} & \textbf{16-shot}
  & \textbf{0-shot} \\
\midrule
CXR
  & Radiology & 5
  & 24.6 (\tiny $\pm$ 2.1)
  & 54.6 (\tiny $\pm$ 1.5)
  & 59.3 (\tiny $\pm$ 1.2)
  & 60.9 (\tiny $\pm$ 0.9)
  & 36.1 (\tiny $\pm$ 0.5) \\
PCam
  & Pathology & 2
  & 58.2 (\tiny $\pm$ 2.4)
  & 78.9 (\tiny $\pm$ 1.3)
  & 82.2 (\tiny $\pm$ 0.8)
  & 81.5 (\tiny $\pm$ 1.0)
  & 70.9 (\tiny $\pm$ 0.8) \\
PathM
  & Pathology & 9
  & 78.2 (\tiny $\pm$ 1.6)
  & 90.9 (\tiny $\pm$ 0.9)
  & 93.5 (\tiny $\pm$ 0.6)
  & 94.3 (\tiny $\pm$ 0.5)
  & 40.6 (\tiny $\pm$ 0.6) \\
BloodM
  & Hematology & 8
  & 61.3 (\tiny $\pm$ 2.3)
  & 78.1 (\tiny $\pm$ 1.4)
  & 86.3 (\tiny $\pm$ 0.9)
  & 89.0 (\tiny $\pm$ 0.7)
  & 26.8 (\tiny $\pm$ 0.4) \\
OrganA
  & CT & 11
  & 49.2 (\tiny $\pm$ 2.5)
  & 63.9 (\tiny $\pm$ 1.6)
  & 71.5 (\tiny $\pm$ 1.1)
  & 76.2 (\tiny $\pm$ 0.8)
  & 21.4 (\tiny $\pm$ 0.3) \\
\bottomrule
\end{tabular}%
}
}
\end{table}



\section{Conclusion and Discussion}
In this work, we investigated a critical paradox in medical foundation models: the significant performance gap between a VLM’s powerful vision encoder and its end-to-end zero-shot diagnostic accuracy. Focusing on MedGemma-4B in the context of dermatology, we systematically investigated three hypotheses for this discrepancy: a train-test distribution mismatch, an over-reliance on language priors, and a fundamental objective misalignment between the vision and language components. To address these challenges, we combined three VLM-fine-tuning-free strategies: providing in-context clinical descriptions, enforcing a DtD reasoning process, and leveraging the vision encoder for candidate selection in a ``Top-5 to Top-1'' framework. Our experiments demonstrate that these interventions, both individually and combined, improve classification accuracy, reduce diagnostic bias, and enhance the model's visual grounding, while retaining the frozen VLM and using a lightweight probe for candidate selection. Experiments on SkinVL, InternVL3, InternVL3.5, and biomed-Qwen support intervention transfer across backbones, while the cross-modality experiments extend the representation and attention analyses. These findings motivate evaluating medical VLMs at both the representation and decision levels, and developing inference strategies that better connect available visual evidence with the final diagnosis.

\textbf{Limitations.} Our detailed analysis centers on MedGemma-4B and dermatology. Additional backbone and modality comparisons broaden the evidence, but the complete pipeline requires evaluation across medical specialties and prospective validation of its clinical utility. Establishing causal visual reliance requires additional controlled interventions.
\bibliographystyle{iclr2027_conference}
\bibliography{reference}

@String(ICLR  = {Int. Conf. Learn. Represent.})

@String(AAAI  = {AAAI})

@String(ICLR  = {ICLR})

@article{Brinkerarticle,
  author = {Brinker, Titus and Hekler, Achim and Enk, Alexander and Klode, Joachim and Hauschild, Axel and Berking, Carola and Schilling, Bastian and Haferkamp, Sebastian and Utikal, Jochen and Kalle, Christof and Fröhling, Stefan and Weichenthal, Michael},
  year = {2019},
  month = {03},
  pages = {148-154},
  title = {A convolutional neural network trained with dermoscopic images performed on par with 145 dermatologists in a clinical melanoma image classification task},
  volume = {111},
  journal = {European Journal of Cancer},
  doi = {10.1016/j.ejca.2019.02.005}
}

@article{Esteva2017DermatologistlevelCO,
  title={Dermatologist-level classification of skin cancer with deep neural networks},
  author={Andre Esteva and Brett Kuprel and Roberto A. Novoa and Justin M. Ko and Susan M. Swetter and Helen M. Blau and Sebastian Thrun},
  journal={Nature},
  year={2017},
  volume={542},
  pages={115-118}
}

@article{liu2020deep,
  title={A deep learning system for differential diagnosis of skin diseases},
  author={Liu, Yuan and Jain, Ayush and Eng, Clara and Way, David H and Lee, Kang and Bui, Peggy and Kanada, Kimberly and de Oliveira Marinho, Guilherme and Gallegos, Jessica and Gabriele, Sara and others},
  journal={Nature medicine},
  volume={26},
  number={6},
  pages={900--908},
  year={2020},
  publisher={Nature Publishing Group}
}

@article{soenksen2021using,
  title={Using deep learning for dermatologist-level detection of suspicious pigmented skin lesions from wide-field images},
  author={Soenksen, Luis R and Kassis, Timothy and Conover, Susan T and Marti-Fuster, Berta and Birkenfeld, Judith S and Tucker-Schwartz, Jason and Naseem, Asif and Stavert, Robert R and Kim, Caroline C and Senna, Maryanne M and others},
  journal={Science Translational Medicine},
  volume={13},
  number={581},
  pages={eabb3652},
  year={2021},
  publisher={American Association for the Advancement of Science}
}

@inproceedings{wang2024achieving,
  title={Achieving reliable and fair skin lesion diagnosis via unsupervised domain adaptation},
  author={Wang, Janet and Zhang, Yunbei and Ding, Zhengming and Hamm, Jihun},
  booktitle={Proceedings of the IEEE/CVF Conference on Computer Vision and Pattern Recognition},
  pages={5157--5166},
  year={2024}
}

@article{yan2025derm1m,
  title={Derm1m: A million-scale vision-language dataset aligned with clinical ontology knowledge for dermatology},
  author={Yan, Siyuan and Hu, Ming and Jiang, Yiwen and Li, Xieji and Fei, Hao and Tschandl, Philipp and Kittler, Harald and Ge, Zongyuan},
  journal={arXiv preprint arXiv:2503.14911},
  year={2025}
}

@article{zeng2025mm,
  title={MM-Skin: Enhancing Dermatology Vision-Language Model with an Image-Text Dataset Derived from Textbooks},
  author={Zeng, Wenqi and Sun, Yuqi and Ma, Chenxi and Tan, Weimin and Yan, Bo},
  journal={arXiv preprint arXiv:2505.06152},
  year={2025}
}

@article{sellergren2025medgemma,
  title={Medgemma technical report},
  author={Sellergren, Andrew and Kazemzadeh, Sahar and Jaroensri, Tiam and Kiraly, Atilla and Traverse, Madeleine and Kohlberger, Timo and Xu, Shawn and Jamil, Fayaz and Hughes, C{\'\i}an and Lau, Charles and others},
  journal={arXiv preprint arXiv:2507.05201},
  year={2025}
}

@article{kim2024transparent,
  title={Transparent medical image AI via an image--text foundation model grounded in medical literature},
  author={Kim, Chanwoo and Gadgil, Soham U and DeGrave, Alex J and Omiye, Jesutofunmi A and Cai, Zhuo Ran and Daneshjou, Roxana and Lee, Su-In},
  journal={Nature medicine},
  volume={30},
  number={4},
  pages={1154--1165},
  year={2024},
  publisher={Nature Publishing Group US New York}
}

@article{zhou2024pre,
  title={Pre-trained multimodal large language model enhances dermatological diagnosis using SkinGPT-4},
  author={Zhou, Juexiao and He, Xiaonan and Sun, Liyuan and Xu, Jiannan and Chen, Xiuying and Chu, Yuetan and Zhou, Longxi and Liao, Xingyu and Zhang, Bin and Afvari, Shawn and others},
  journal={Nature Communications},
  volume={15},
  number={1},
  pages={5649},
  year={2024},
  publisher={Nature Publishing Group UK London}
}

@article{team2025gemma,
  title={Gemma 3 technical report},
  author={Team, Gemma and Kamath, Aishwarya and Ferret, Johan and Pathak, Shreya and Vieillard, Nino and Merhej, Ramona and Perrin, Sarah and Matejovicova, Tatiana and Ram{\'e}, Alexandre and Rivi{\`e}re, Morgane and others},
  journal={arXiv preprint arXiv:2503.19786},
  year={2025}
}

@article{hu2022lora,
  title={Lora: Low-rank adaptation of large language models.},
  author={Hu, Edward J and Shen, Yelong and Wallis, Phillip and Allen-Zhu, Zeyuan and Li, Yuanzhi and Wang, Shean and Wang, Lu and Chen, Weizhu and others},
  journal={ICLR},
  volume={1},
  number={2},
  pages={3},
  year={2022}
}

@article{Kawahara2018-7pt,
author = {Kawahara, Jeremy and Daneshvar, Sara and Argenziano, Giuseppe and Hamarneh, Ghassan},
doi = {10.1109/JBHI.2018.2824327},
issn = {2168-2194},
journal = {IEEE Journal of Biomedical and Health Informatics},
month = {mar},
number = {2},
pages = {538--546},
publisher = {IEEE},
title = {Seven-point checklist and skin lesion classification using multitask multimodal neural nets},
volume = {23},
year = {2019}
}

@article{wang2025eskinhealth,
  title={eSkinHealth: A Multimodal Dataset for Neglected Tropical Skin Diseases},
  author={Wang, Janet and Hu, Xin and Zhang, Yunbei and Almamy, Diabate and Bamba, Vagamon and Koffi, Konan Amos S{\'e}bastien and Aubin, Yao Koffi and Ding, Zhengming and Hamm, Jihun and Yotsu, Rie R},
  journal={arXiv preprint arXiv:2508.18608},
  year={2025}
}

@article{wang2025doctor,
  title={Doctor Approved: Generating Medically Accurate Skin Disease Images through AI-Expert Feedback},
  author={Wang, Janet and Zhang, Yunbei and Ding, Zhengming and Hamm, Jihun},
  journal={arXiv preprint arXiv:2506.12323},
  year={2025}
}

@inproceedings{groh2021evaluating,
  title={Evaluating deep neural networks trained on clinical images in dermatology with the fitzpatrick 17k dataset},
  author={Groh, Matthew and Harris, Caleb and Soenksen, Luis and Lau, Felix and Han, Rachel and Kim, Aerin and Koochek, Arash and Badri, Omar},
  booktitle={Proceedings of the IEEE/CVF conference on computer vision and pattern recognition},
  pages={1820--1828},
  year={2021}
}

@article{pacheco2020pad,
  title={PAD-UFES-20: A skin lesion dataset composed of patient data and clinical images collected from smartphones},
  author={Pacheco, Andre GC and Lima, Gustavo R and Salomao, Amanda S and Krohling, Breno and Biral, Igor P and De Angelo, Gabriel G and Alves Jr, F{\'a}bio CR and Esgario, Jos{\'e} GM and Simora, Alana C and Castro, Pedro BC and others},
  journal={Data in brief},
  volume={32},
  pages={106221},
  year={2020},
  publisher={Elsevier}
}

@article{yang2019self,
  title={Self-paced balance learning for clinical skin disease recognition},
  author={Yang, Jufeng and Wu, Xiaoping and Liang, Jie and Sun, Xiaoxiao and Cheng, Ming-Ming and Rosin, Paul L and Wang, Liang},
  journal={IEEE transactions on neural networks and learning systems},
  volume={31},
  number={8},
  pages={2832--2846},
  year={2019},
  publisher={IEEE}
}

@inproceedings{zhai2023sigmoid,
  title={Sigmoid loss for language image pre-training},
  author={Zhai, Xiaohua and Mustafa, Basil and Kolesnikov, Alexander and Beyer, Lucas},
  booktitle={Proceedings of the IEEE/CVF international conference on computer vision},
  pages={11975--11986},
  year={2023}
}

@article{yang2025look,
  title={Look-Back: Implicit Visual Re-focusing in MLLM Reasoning},
  author={Yang, Shuo and Niu, Yuwei and Liu, Yuyang and Ye, Yang and Lin, Bin and Yuan, Li},
  journal={arXiv preprint arXiv:2507.03019},
  year={2025}
}

@inproceedings{tong2024eyes,
  title={Eyes wide shut? exploring the visual shortcomings of multimodal llms},
  author={Tong, Shengbang and Liu, Zhuang and Zhai, Yuexiang and Ma, Yi and LeCun, Yann and Xie, Saining},
  booktitle={Proceedings of the IEEE/CVF Conference on Computer Vision and Pattern Recognition},
  pages={9568--9578},
  year={2024}
}

@article{yan2025multimodal,
  title={A multimodal vision foundation model for clinical dermatology},
  author={Yan, Siyuan and Yu, Zhen and Primiero, Clare and Vico-Alonso, Cristina and Wang, Zhonghua and Yang, Litao and Tschandl, Philipp and Hu, Ming and Ju, Lie and Tan, Gin and others},
  journal={Nature Medicine},
  pages={1--12},
  year={2025},
  publisher={Nature Publishing Group US New York}
}

@inproceedings{radford2021learning,
  title={Learning transferable visual models from natural language supervision},
  author={Radford, Alec and Kim, Jong Wook and Hallacy, Chris and Ramesh, Aditya and Goh, Gabriel and Agarwal, Sandhini and Sastry, Girish and Askell, Amanda and Mishkin, Pamela and Clark, Jack and others},
  booktitle={International conference on machine learning},
  pages={8748--8763},
  year={2021},
  organization={PmLR}
}

@article{zhai2023investigating,
  title={Investigating the catastrophic forgetting in multimodal large language models},
  author={Zhai, Yuexiang and Tong, Shengbang and Li, Xiao and Cai, Mu and Qu, Qing and Lee, Yong Jae and Ma, Yi},
  journal={arXiv preprint arXiv:2309.10313},
  year={2023}
}

@article{hu2024visual,
  title={Visual sketchpad: Sketching as a visual chain of thought for multimodal language models},
  author={Hu, Yushi and Shi, Weijia and Fu, Xingyu and Roth, Dan and Ostendorf, Mari and Zettlemoyer, Luke and Smith, Noah A and Krishna, Ranjay},
  journal={Advances in Neural Information Processing Systems},
  volume={37},
  pages={139348--139379},
  year={2024}
}

@article{zhang2024generalist,
  title={A generalist vision--language foundation model for diverse biomedical tasks},
  author={Zhang, Kai and Zhou, Rong and Adhikarla, Eashan and Yan, Zhiling and Liu, Yixin and Yu, Jun and Liu, Zhengliang and Chen, Xun and Davison, Brian D and Ren, Hui and others},
  journal={Nature Medicine},
  volume={30},
  number={11},
  pages={3129--3141},
  year={2024},
  publisher={Nature Publishing Group US New York}
}

@article{touvron2023llama,
  title={Llama 2: Open foundation and fine-tuned chat models},
  author={Touvron, Hugo and Martin, Louis and Stone, Kevin and Albert, Peter and Almahairi, Amjad and Babaei, Yasmine and Bashlykov, Nikolay and Batra, Soumya and Bhargava, Prajjwal and Bhosale, Shruti and others},
  journal={arXiv preprint arXiv:2307.09288},
  year={2023}
}

@inproceedings{lin2023pmc,
  title={Pmc-clip: Contrastive language-image pre-training using biomedical documents},
  author={Lin, Weixiong and Zhao, Ziheng and Zhang, Xiaoman and Wu, Chaoyi and Zhang, Ya and Wang, Yanfeng and Xie, Weidi},
  booktitle={International Conference on Medical Image Computing and Computer-Assisted Intervention},
  pages={525--536},
  year={2023},
  organization={Springer}
}

@article{xing2025re,
  title={Re-Align: Aligning Vision Language Models via Retrieval-Augmented Direct Preference Optimization},
  author={Xing, Shuo and Wang, Yuping and Li, Peiran and Bai, Ruizheng and Wang, Yueqi and Hu, Chan-wei and Qian, Chengxuan and Yao, Huaxiu and Tu, Zhengzhong},
  journal={arXiv preprint arXiv:2502.13146},
  year={2025}
}

@inproceedings{delbrouck2025automated,
  title={Automated structured radiology report generation},
  author={Delbrouck, Jean-Benoit and Xu, Justin and Moll, Johannes and Thomas, Alois and Chen, Zhihong and Ostmeier, Sophie and Azhar, Asfandyar and Li, Kelvin Zhenghao and Johnston, Andrew and Bluethgen, Christian and others},
  booktitle={Proceedings of the 63rd Annual Meeting of the Association for Computational Linguistics (Volume 1: Long Papers)},
  pages={26813--26829},
  year={2025}
}

@article{holland2025specialized,
  title={Specialized curricula for training vision language models in retinal image analysis},
  author={Holland, Robbie and Taylor, Thomas RP and Holmes, Christopher and Riedl, Sophie and Mai, Julia and Patsiamanidi, Maria and Mitsopoulou, Dimitra and Hager, Paul and M{\"u}ller, Philip and Paetzold, Johannes C and others},
  journal={NPJ Digital Medicine},
  volume={8},
  number={1},
  pages={532},
  year={2025},
  publisher={Nature Publishing Group UK London}
}

@article{dettmers2023qlora,
  title={Qlora: Efficient finetuning of quantized llms},
  author={Dettmers, Tim and Pagnoni, Artidoro and Holtzman, Ari and Zettlemoyer, Luke},
  journal={Advances in neural information processing systems},
  volume={36},
  pages={10088--10115},
  year={2023}
}

@inproceedings{irvin2019chexpert,
  title={Chexpert: A large chest radiograph dataset with uncertainty labels and expert comparison},
  author={Irvin, Jeremy and Rajpurkar, Pranav and Ko, Michael and Yu, Yifan and Ciurea-Ilcus, Silviana and Chute, Chris and Marklund, Henrik and Haghgoo, Behzad and Ball, Robyn and Shpanskaya, Katie and others},
  booktitle={Proceedings of the AAAI conference on artificial intelligence},
  volume={33},
  number={01},
  pages={590--597},
  year={2019}
}

@ARTICLE{Veeling2018-qh,
  title         = "Rotation Equivariant {CNNs} for Digital Pathology",
  author        = "Veeling, Bastiaan S and Linmans, Jasper and Winkens, Jim and
                   Cohen, Taco and Welling, Max",
  month         =  jun,
  year          =  2018,
  archivePrefix = "arXiv",
  primaryClass  = "cs.CV",
  eprint        = "1806.03962"
}

@article{medmnistv2,
    title={MedMNIST v2-A large-scale lightweight benchmark for 2D and 3D biomedical image classification},
    author={Yang, Jiancheng and Shi, Rui and Wei, Donglai and Liu, Zequan and Zhao, Lin and Ke, Bilian and Pfister, Hanspeter and Ni, Bingbing},
    journal={Scientific Data},
    volume={10},
    number={1},
    pages={41},
    year={2023},
    publisher={Nature Publishing Group UK London}
}

@article{ferber2024context,
  title={In-context learning enables multimodal large language models to classify cancer pathology images},
  author={Ferber, Dyke and W{\"o}lflein, Georg and Wiest, Isabella C and Ligero, Marta and Sainath, Srividhya and Ghaffari Laleh, Narmin and El Nahhas, Omar SM and M{\"u}ller-Franzes, Gustav and J{\"a}ger, Dirk and Truhn, Daniel and others},
  journal={Nature Communications},
  volume={15},
  number={1},
  pages={10104},
  year={2024},
  publisher={Nature Publishing Group UK London}
}

@inproceedings{cheng-etal-2025-domain,
  title = {On Domain-Adaptive Post-Training for Multimodal Large Language Models},
  author = {Cheng, Daixuan and Huang, Shaohan and Zhu, Ziyu and Zhang, Xintong and Zhao, Wayne Xin and Luan, Zhongzhi and Dai, Bo and Zhang, Zhenliang},
  editor = {Christodoulopoulos, Christos and Chakraborty, Tanmoy and Rose, Carolyn and Peng, Violet},
  booktitle = {Findings of the Association for Computational Linguistics: EMNLP 2025},
  month = nov,
  year = {2025},
  address = {Suzhou, China},
  publisher = {Association for Computational Linguistics},
  url = {https://aclanthology.org/2025.findings-emnlp.17/},
  doi = {10.18653/v1/2025.findings-emnlp.17},
  pages = {274--296},
  ISBN = {979-8-89176-335-7}
}

@misc{zhu2025internvl3exploringadvancedtraining,
  title = {InternVL3: Exploring Advanced Training and Test-Time Recipes for Open-Source Multimodal Models},
  author = {Jinguo Zhu and Weiyun Wang and Zhe Chen and Zhaoyang Liu and Shenglong Ye and Lixin Gu and Hao Tian and Yuchen Duan and Weijie Su and Jie Shao and Zhangwei Gao and Erfei Cui and Xuehui Wang and Yue Cao and Yangzhou Liu and Xingguang Wei and Hongjie Zhang and Haomin Wang and Weiye Xu and Hao Li and Jiahao Wang and Nianchen Deng and Songze Li and Yinan He and Tan Jiang and Jiapeng Luo and Yi Wang and Conghui He and Botian Shi and Xingcheng Zhang and Wenqi Shao and Junjun He and Yingtong Xiong and Wenwen Qu and Peng Sun and Penglong Jiao and Han Lv and Lijun Wu and Kaipeng Zhang and Huipeng Deng and Jiaye Ge and Kai Chen and Limin Wang and Min Dou and Lewei Lu and Xizhou Zhu and Tong Lu and Dahua Lin and Yu Qiao and Jifeng Dai and Wenhai Wang},
  year = {2025},
  eprint = {2504.10479},
  archivePrefix = {arXiv},
  primaryClass = {cs.CV},
  url = {https://arxiv.org/abs/2504.10479}
}

@misc{wang2025internvl35advancingopensourcemultimodal,
  title = {InternVL3.5: Advancing Open-Source Multimodal Models in Versatility, Reasoning, and Efficiency},
  author = {Weiyun Wang and Zhangwei Gao and Lixin Gu and Hengjun Pu and Long Cui and Xingguang Wei and Zhaoyang Liu and Linglin Jing and Shenglong Ye and Jie Shao and Zhaokai Wang and Zhe Chen and Hongjie Zhang and Ganlin Yang and Haomin Wang and Qi Wei and Jinhui Yin and Wenhao Li and Erfei Cui and Guanzhou Chen and Zichen Ding and Changyao Tian and Zhenyu Wu and Jingjing Xie and Zehao Li and Bowen Yang and Yuchen Duan and Xuehui Wang and Zhi Hou and Haoran Hao and Tianyi Zhang and Songze Li and Xiangyu Zhao and Haodong Duan and Nianchen Deng and Bin Fu and Yinan He and Yi Wang and Conghui He and Botian Shi and Junjun He and Yingtong Xiong and Han Lv and Lijun Wu and Wenqi Shao and Kaipeng Zhang and Huipeng Deng and Biqing Qi and Jiaye Ge and Qipeng Guo and Wenwei Zhang and Songyang Zhang and Maosong Cao and Junyao Lin and Kexian Tang and Jianfei Gao and Haian Huang and Yuzhe Gu and Chengqi Lyu and Huanze Tang and Rui Wang and Haijun Lv and Wanli Ouyang and Limin Wang and Min Dou and Xizhou Zhu and Tong Lu and Dahua Lin and Jifeng Dai and Weijie Su and Bowen Zhou and Kai Chen and Yu Qiao and Wenhai Wang and Gen Luo},
  year = {2025},
  eprint = {2508.18265},
  archivePrefix = {arXiv},
  primaryClass = {cs.CV},
  url = {https://arxiv.org/abs/2508.18265}
}

@misc{zhang2025biomedclipmultimodalbiomedicalfoundation,
  title = {BiomedCLIP: a multimodal biomedical foundation model pretrained from fifteen million scientific image-text pairs},
  author = {Sheng Zhang and Yanbo Xu and Naoto Usuyama and Hanwen Xu and Jaspreet Bagga and Robert Tinn and Sam Preston and Rajesh Rao and Mu Wei and Naveen Valluri and Cliff Wong and Andrea Tupini and Yu Wang and Matt Mazzola and Swadheen Shukla and Lars Liden and Jianfeng Gao and Angela Crabtree and Brian Piening and Carlo Bifulco and Matthew P. Lungren and Tristan Naumann and Sheng Wang and Hoifung Poon},
  year = {2025},
  eprint = {2303.00915},
  archivePrefix = {arXiv},
  primaryClass = {cs.CV},
  url = {https://arxiv.org/abs/2303.00915}
}

@article{lau2018dataset,
  title = {A dataset of clinically generated visual questions and answers about radiology images},
  author = {Lau, Jason J and Gayen, Soumya and Ben Abacha, Asma and Demner-Fushman, Dina},
  journal = {Scientific data},
  volume = {5},
  number = {1},
  pages = {180251},
  year = {2018},
  publisher = {Nature Publishing Group}
}

@inproceedings{zhang2025dpcore,
  title={{DPCore}: Dynamic Prompt Coreset for Continual Test-Time Adaptation},
  author={Zhang, Yunbei and Mehra, Akshay and Niu, Shuaicheng and Hamm, Jihun},
  booktitle={International Conference on Machine Learning},
  pages={75757--75778},
  year={2025},
  organization={PMLR}
}

@inproceedings{zhang2025ot,
  title={{OT-VP}: Optimal Transport-Guided Visual Prompting for Test-Time Adaptation},
  author={Zhang, Yunbei and Mehra, Akshay and Hamm, Jihun},
  booktitle={2025 IEEE/CVF Winter Conference on Applications of Computer Vision (WACV)},
  pages={1122--1132},
  year={2025},
  organization={IEEE}
}

@article{zhang2026adapting,
  title={Adapting in the Dark: Efficient and Stable Test-Time Adaptation for Black-Box Models},
  author={Zhang, Yunbei and Niu, Shuaicheng and Cai, Chengyi and Liu, Feng and Hamm, Jihun},
  journal={arXiv preprint arXiv:2604.15609},
  year={2026}
}

@inproceedings{zhang2026prime,
  title={Prime Once, then Reprogram Locally: An Efficient Alternative to Black-Box Service Model Adaptation},
  author={Zhang, Yunbei and Cai, Chengyi and Liu, Feng and Hamm, Jihun},
  booktitle={Proceedings of the IEEE/CVF Conference on Computer Vision and Pattern Recognition},
  pages={6176--6187},
  year={2026}
}

\clearpage
\appendix

\begin{center}
    \huge \textbf{\texttt{Appendix}}
\end{center}

We provide additional baseline and prompt-control results, evaluations on other backbones and VQA-RAD, and extended confusion-matrix and attention analyses. The appendix also contains qualitative examples, evaluation prompts, implementation details, shot ablations, SkinVL comparisons, dataset statistics and clinical descriptions, and a discussion of limitations.
\section{Additional Baselines and Controls}
\label{sec:additional_baselines}
\subsection{Zero-Label Encoder Comparison}
\label{sec:zero_label_protocol}
We evaluate MedSigLIP through image--text similarity using the same candidate class names provided to MedGemma. This evaluation uses no labeled target images, parameter updates, or fitted classifier. Table~\ref{tab:zero_label_full} also reports BiomedCLIP~\citep{zhang2025biomedclipmultimodalbiomedicalfoundation} and DermLIP~\citep{yan2025derm1m} as additional contrastive baselines. MedSigLIP exceeds MedGemma on each of the four datasets, with an average accuracy difference of 10.26 percentage points.

\begin{table}[!ht]
\centering\small
\setlength{\tabcolsep}{7pt}
\caption{\textbf{Zero-label encoder baselines outperform MedGemma in most comparisons.} Top-1 accuracy (\%) without fitting a target-task classifier. MedSigLIP outperforms MedGemma on all four datasets.}
\label{tab:zero_label_full}
\begin{tabular}{@{}lrrrr@{}}
\toprule
Dataset & MedGemma & MedSigLIP & BiomedCLIP & DermLIP \\
\midrule
Derm7pt & 16.46 & 24.18 & 24.55 & 24.03 \\
eSkinHealth & 13.35 & 26.53 & 28.22 & 29.28 \\
Fitzpatrick17k & 16.73 & 28.37 & 25.71 & 31.55 \\
PAD-UFES-20 & 46.91 & 55.42 & 46.39 & 53.06 \\
\bottomrule
\end{tabular}
\end{table}
\subsection{Multimodal Image In-Context Learning}
\label{sec:image_icl}
The image-ICL baseline supplies MedGemma with labeled image--class pairs as demonstrations. Table~\ref{tab:image_icl_full} compares this 8-shot setting with the zero-shot VLM, the 8-shot linear probe, and encoder-assisted reranking. Image ICL improves accuracy on all four datasets, by 3.61 percentage points on average. The 8-shot LP remains stronger in each comparison. Reranking gives the highest accuracy on Derm7pt, eSkinHealth, and PAD-UFES-20; on Fitzpatrick17k its accuracy is 0.61 points below LP. The full 1-shot and 8-shot pipeline comparison, including the original reported variability, is retained in Table~\ref{tab:pipeline_performance_detailed}.

\begin{table}[!ht]
\centering\small
\setlength{\tabcolsep}{7pt}
\caption{\textbf{Image ICL improves the VLM baseline while leaving a gap to LP.} Top-1 accuracy (\%). The image-ICL, LP, and reranking columns use the 8-shot settings.}
\label{tab:image_icl_full}
\begin{tabular}{@{}lrrrr@{}}
\toprule
Dataset & VLM 0-shot & Image ICL & LP & Reranking \\
\midrule
Derm7pt & 16.46 & 19.44 & 28.66 & 35.56 \\
eSkinHealth & 13.35 & 18.15 & 38.80 & 42.05 \\
Fitzpatrick17k & 16.73 & 20.72 & 38.43 & 37.82 \\
PAD-UFES-20 & 46.91 & 49.58 & 53.33 & 80.31 \\
\bottomrule
\end{tabular}
\end{table}

\begin{table}[!ht]
    \centering
    \caption{Effectiveness of the proposed two-stage inference pipeline. Comparison of the \textbf{Baseline VLM} (MedGemma zero-shot), the standalone \textbf{Linear Probe} (MedSigLIP), and \textbf{Our Pipeline} which uses the Linear Probe to filter candidates (Top-5) for the VLM (Top-1). All accuracy values are Top-1 (\%). Ablation study on different number of shots can be found in Appendix \ref{ablation_shots}.}
    \label{tab:pipeline_performance_detailed}
    \tiny
    \resizebox{\linewidth}{!}{
    \begin{tabular}{l | c | cc | cc}
        \toprule
         & \textbf{Baseline VLM} & \multicolumn{2}{c|}{\textbf{Linear Probe}} & \multicolumn{2}{c}{\textbf{Our Pipeline}} \\
         \textbf{Dataset}& \textit{Zero-Shot} & \multicolumn{2}{c|}{\textit{(Vision Encoder Only)}} & \multicolumn{2}{c}{\textit{(LP Top-5 $\rightarrow$ VLM Top-1)}} \\
        \cmidrule(lr){2-2} \cmidrule(lr){3-4} \cmidrule(lr){5-6}
         & \textbf{0-shot} & \textbf{1-shot} & \textbf{8-shot} & \textbf{1-shot} & \textbf{8-shot} \\
        \midrule
        Derm7pt & 16.46 (\tiny $\pm$ 0.01) & 12.96 (\tiny $\pm$ 1.75) & 28.66 (\tiny $\pm$ 1.38) & 20.02 (\tiny $\pm$ 2.39) & 35.56 (\tiny $\pm$ 0.99) \\
        eSkinHealth & 13.35 (\tiny $\pm$ 0.23) & 23.51 (\tiny $\pm$ 1.91) & 38.80 (\tiny $\pm$ 1.06) & 30.98 (\tiny $\pm$ 2.85) & 42.05 (\tiny $\pm$ 1.48) \\
        Fitzpatrick17k & 16.73 (\tiny $\pm$ 0.19) & 17.63 (\tiny $\pm$ 3.13) & 38.43 (\tiny $\pm$ 1.03) & 25.35 (\tiny $\pm$ 2.41) & 37.82 (\tiny $\pm$ 1.37) \\
        SD-260 & 10.81 (\tiny $\pm$ 0.13) & 11.48 (\tiny $\pm$ 2.27) & 37.12 (\tiny $\pm$ 1.46) & 22.31 (\tiny $\pm$ 1.95) & 42.33 (\tiny $\pm$ 0.92) \\
        PAD-UFES-20 & 46.91 (\tiny $\pm$ 0.20) & 34.85 (\tiny $\pm$ 1.83) & 53.33 (\tiny $\pm$ 1.34) & 58.33 (\tiny $\pm$ 2.24) & 80.31 (\tiny $\pm$ 1.23) \\
        \bottomrule
    \end{tabular}
    }
\end{table}
\subsection{Clinical-Description Controls}
\label{sec:context_controls}
We compare class names alone with three types of additional text: class-aligned clinical descriptions, length-matched generic dermatology text without class-discriminative information, and descriptions assigned to incorrect classes. Table~\ref{tab:context_controls} shows that correctly aligned descriptions perform best among these four prompt-only settings on both datasets. Generic text does not improve over names alone, and shuffled descriptions reduce accuracy. The comparison supports a role for the alignment of clinical information with the candidate labels.

\begin{table}[!ht]
\centering\small
\setlength{\tabcolsep}{7pt}
\caption{\textbf{Class-aligned clinical information improves over prompt-length controls.} Top-1 accuracy (\%). The first four rows use no labeled target images; the last two include an 8-shot probe.}
\label{tab:context_controls}
\begin{tabular}{@{}lrr@{}}
\toprule
Setting & Fitzpatrick17k & PAD-UFES-20 \\
\midrule
Names only & 16.73 & 46.22 \\
Class-aligned descriptions & 24.01 & 52.23 \\
Length-matched generic text & 16.68 & 36.36 \\
Shuffled descriptions & 12.71 & 15.64 \\
\midrule
Top-5-to-Top-1 & 28.32 & 75.23 \\
All combined & 40.97 & 84.68 \\
\bottomrule
\end{tabular}
\end{table}
\subsection{Full-Data Classification Comparison}
\begin{table}[!ht]
\centering

\caption{Classification performance across five dermatology datasets. The table compares the performance for MedGemma-4B under three conditions: zero-shot inference, full-data fine-tuning, and a linear probe on its vision encoder (MedSigLIP). Performance is compared against a linear probe on a competitive vision encoder, PanDerm. Linear probing on both vision encoders significantly outperforms the zero-shot with the VLM, even after fine-tuning. Results are averaged over five runs.}
\vspace{-1mm}
\label{Table_1}
\scriptsize
\resizebox{\textwidth}{!}{%
\begin{tabular}{l|r|r|r|cc|cc|cc|cc}
\toprule
\multicolumn{4}{c|}{\textbf{Data Details}} &
\multicolumn{6}{c|}{\cellcolor{MedGemma}\textbf{MedGemma-4B}} &
\multicolumn{2}{c}{\cellcolor{PanDerm}\textbf{PanDerm}} \\
\midrule
\multirow{2}{*}{\textbf{Datasets}} &
\multirow{2}{*}{\textbf{\#Class}} &
\multirow{2}{*}{\textbf{\#Train}} &
\multirow{2}{*}{\textbf{\#Test}} &
\multicolumn{2}{c|}{\textbf{Zero-Shot}} &
\multicolumn{2}{c|}{\textbf{Fine-tune}} &
\multicolumn{2}{c|}{\textbf{Linear Probe}} &
\multicolumn{2}{c}{\textbf{Linear Probe}} \\
\cmidrule(lr){5-12}
& & & &
\textbf{ACC} & \textbf{F1} &
\textbf{ACC} & \textbf{F1} &
\textbf{ACC} & \textbf{F1} &
\textbf{ACC} & \textbf{F1} \\
\midrule
Derm7pt     & 14  & 413   & 395   & 16.46 (\tiny $\pm$ 0.02) & 17.77 & 39.50 (\tiny $\pm$ 2.15) & 22.49 & 57.22 (\tiny $\pm$ 0.88) & 30.07 & 58.99 (\tiny $\pm$ 0.76) & 57.93 \\
eSkinHealth     & 24  & 2,714 & 2,676 & 13.35 (\tiny $\pm$ 0.15) & 16.33 & 60.26 (\tiny $\pm$ 1.42) & 56.17 & 65.35 (\tiny $\pm$ 0.54) & 42.51 & 60.64 (\tiny $\pm$ 0.61) & 59.83 \\
Fitzpatrick17k  & 20  & 3,100 & 3,100 & 16.73 (\tiny $\pm$ 0.08) & 11.13 & 47.32 (\tiny $\pm$ 1.10) & 45.16 & 66.32 (\tiny $\pm$ 0.45) & 66.90 & 64.87 (\tiny $\pm$ 0.39) & 64.88 \\
SD-260          & 260 & 10,362 & 10,238 & 10.81 (\tiny $\pm$ 0.05) & 7.94 & 39.45 (\tiny $\pm$ 0.95) & 34.63 & 75.35 (\tiny $\pm$ 0.32) & 60.10 & 72.08 (\tiny $\pm$ 0.28) & 60.24 \\
PAD-UFES-20     & 6   & 1,134  & 1,164  & 46.91 (\tiny $\pm$ 0.12) & 32.98 & 66.75 (\tiny $\pm$ 1.85) & 61.53 & 76.03 (\tiny $\pm$ 0.91) & 66.22 & 74.21 (\tiny $\pm$ 0.84) & 64.98 \\
\bottomrule
\end{tabular}%
}
\vspace{-1mm}
\end{table}
\section{Additional Backbone Evaluation}
\label{sec:additional_models}
We evaluate the general-purpose InternVL3-4B~\citep{zhu2025internvl3exploringadvancedtraining} and InternVL3.5-4B~\citep{wang2025internvl35advancingopensourcemultimodal}, together with the medical biomed-Qwen2.5-VL-3B~\citep{cheng-etal-2025-domain}, on Derm7pt and Fitzpatrick17k. The InternVL3.5 evaluation uses the same settings as InternVL3. Table~\ref{tab:additional_backbones_full} retains all six model--dataset comparisons. DtD improves direct-answer accuracy in every comparison, and Top-5-to-Top-1 gives the strongest result among the evaluated inference settings. These results test transfer of the interventions across backbones.

\begin{table}[!ht]
\centering\small
\setlength{\tabcolsep}{5pt}
\caption{\textbf{DtD and candidate-assisted inference improve all six added backbone--dataset pairs.} Top-1 accuracy (\%). Top-5-to-Top-1 uses the 8-shot configuration.}
\label{tab:additional_backbones_full}
\begin{tabular}{@{}llrrr@{}}
\toprule
Model & Dataset & DA & DtD & Top-5-to-Top-1 \\
\midrule
InternVL3-4B & Derm7pt & 5.68 & 6.20 & 14.83 \\
& Fitzpatrick17k & 10.10 & 12.23 & 19.70 \\
InternVL3.5-4B & Derm7pt & 7.21 & 7.85 & 18.32 \\
& Fitzpatrick17k & 9.25 & 13.15 & 22.16 \\
biomed-Qwen2.5-VL-3B & Derm7pt & 20.93 & 38.50 & 41.24 \\
& Fitzpatrick17k & 12.14 & 19.96 & 22.35 \\
\bottomrule
\end{tabular}
\end{table}
\section{Cross-Domain VQA after Task-Specific Fine-Tuning}
\label{sec:vqa_transfer}
We evaluate the base MedGemma-4B and its Fitzpatrick17k QLoRA checkpoint on VQA-RAD~\citep{lau2018dataset}. The adapted checkpoint is trained on the full Fitzpatrick17k training set and improves in-domain classification from 16.73\% to 47.32\%. Both checkpoints are evaluated on the same 451 VQA-RAD questions with identical prompts and greedy decoding. Table~\ref{tab:vqa_transfer} shows decreases of 5.04 points in closed-ended accuracy, 15.00 points in open-ended accuracy, and 22.53 points in token recall after adaptation. This comparison measures the cross-domain medical question-answering trade-off associated with task-specific fine-tuning.

\begin{table}[!ht]
\centering\small
\setlength{\tabcolsep}{8pt}
\caption{\textbf{Dermatology adaptation improves in-domain classification but reduces cross-domain VQA performance.} Accuracy and token recall (\%) on VQA-RAD; higher is better.}
\label{tab:vqa_transfer}
\begin{tabular}{@{}lrrr@{}}
\toprule
Model & Closed-ended & Open-ended & Token recall \\
\midrule
Base MedGemma-4B & 70.92 & 37.50 & 49.96 \\
Fitzpatrick17k QLoRA & 65.88 & 22.50 & 27.43 \\
\bottomrule
\end{tabular}
\end{table}
\section{Full Confusion-Matrix Comparison}
\label{sec:full_confusion}
\begin{figure}[!ht]
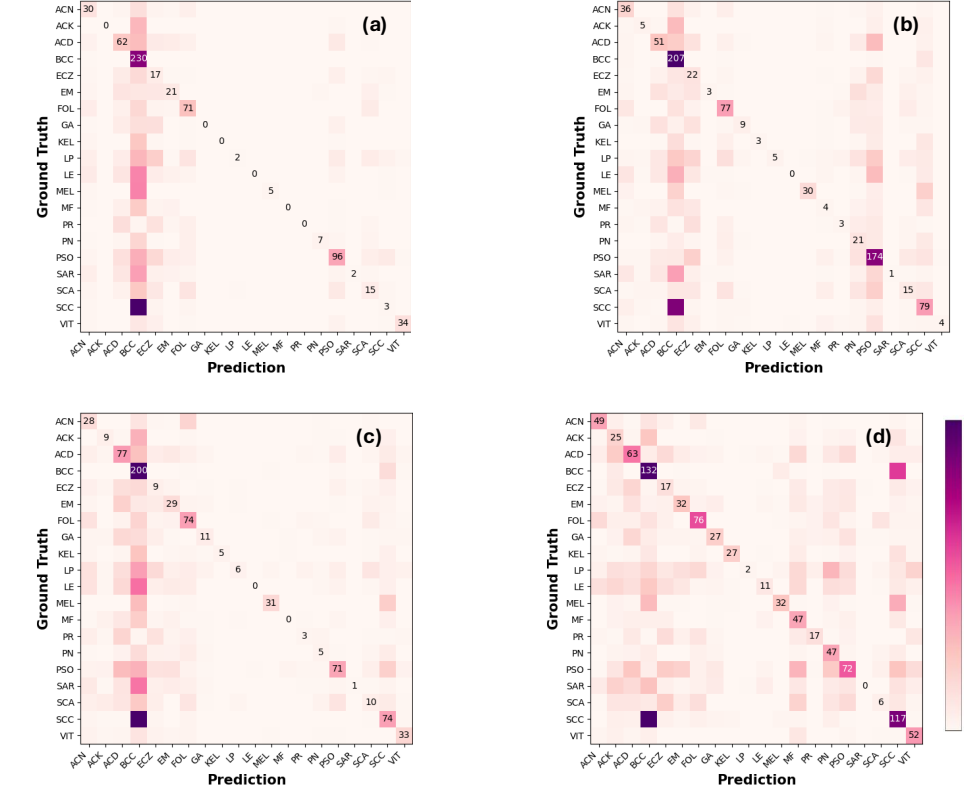

\centering
\begin{subfigure}[t]{0.49\linewidth}\centering
\includegraphics[height=2.05in,viewport=0 0 225 216,clip]{figures/confusion_matrix.pdf}
\end{subfigure}\hfill
\begin{subfigure}[t]{0.49\linewidth}\centering
\includegraphics[height=2.05in,viewport=222 0 447 216,clip]{figures/confusion_matrix.pdf}
\end{subfigure}
\par\medskip
\begin{subfigure}[t]{0.49\linewidth}\centering
\includegraphics[height=2.05in,viewport=443 0 668 216,clip]{figures/confusion_matrix.pdf}
\end{subfigure}\hfill
\begin{subfigure}[t]{0.49\linewidth}\centering
\includegraphics[height=2.05in,viewport=665 0 921.75 216,clip]{figures/confusion_matrix.pdf}
\end{subfigure}
\caption{\textbf{Prompting and candidate selection change diagnostic confusion.} Original Fitzpatrick17k matrices for (a) direct answer, (b) clinical descriptions, (c) DtD, and (d) Top-5-to-Top-1 with a 2-shot probe. Panels retain the original counts and are enlarged for readability. Class abbreviations appear in Table~\ref{tab:fitz_distribution}.}
\label{fig:confusion_matrix_full}
\end{figure}

Figure~\ref{fig:confusion_matrix_full} presents the full confusion matrices for the four inference settings compared in our dermatology analysis.
\section{Attention Heatmap Construction}
\label{sec:heatmap_construction}
To examine how DtD changes attention to the lesion area, we generate and analyze attention heatmaps. Here, we define two metrics:
\textbf{Label-Token Heatmap} ($\mathbf m_{\text{label}}$): This measures the average attention the model pays to the input image tokens specifically when generating the words of the final diagnosis (e.g., ``basal cell carcinoma''). It helps us see what the model attends to at the moment of decision.
\textbf{All-Tokens Heatmap} ($\mathbf m_{\text{all}}$): This measures the average attention paid to the image tokens across the entire generated response, including both the description and the diagnosis.

\begin{figure*}[!ht]
    \centering
    \includegraphics[width=0.9\linewidth]{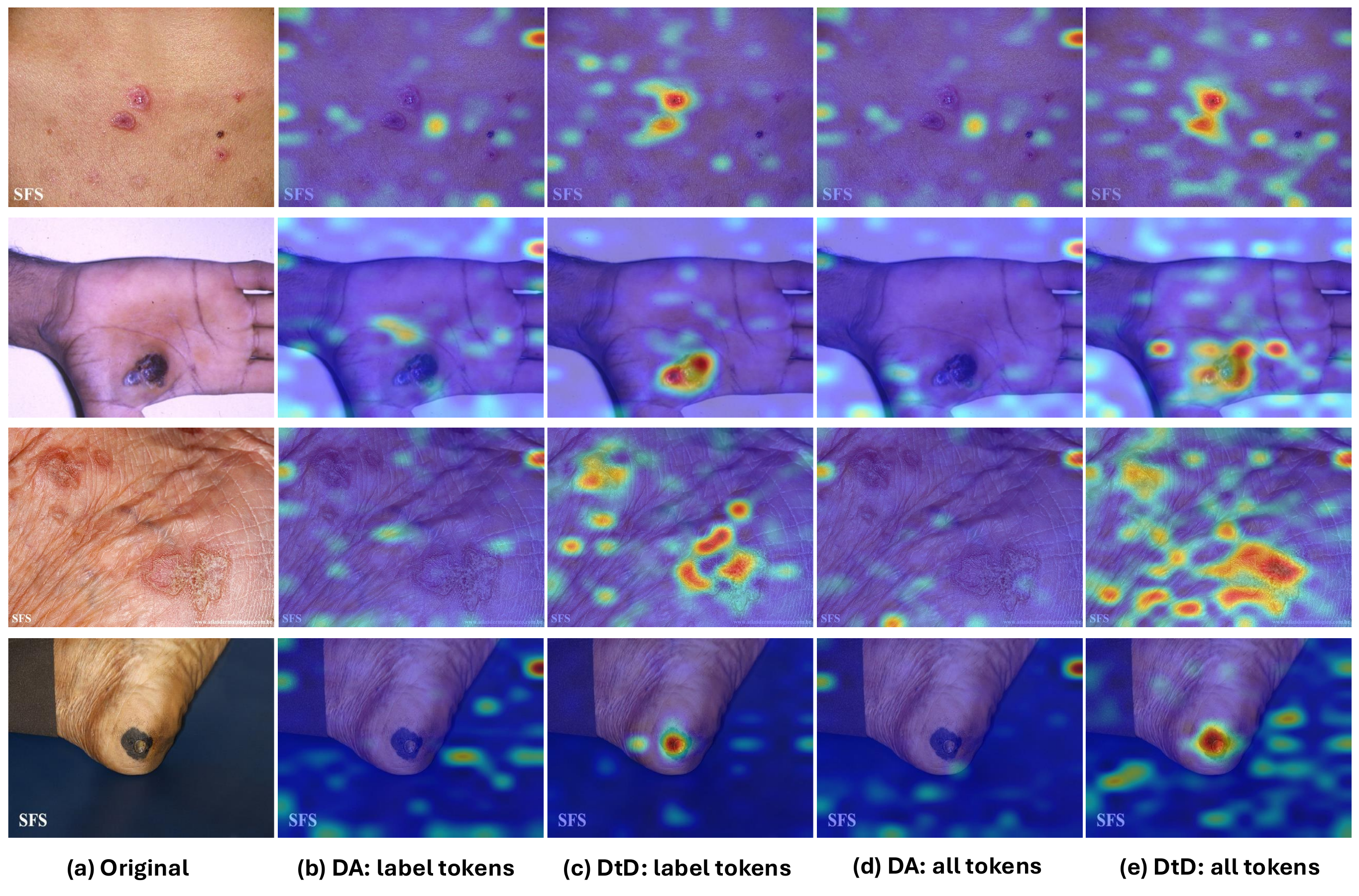}
    \vspace{-3mm}
    \caption{Attention heatmaps. Columns show (a) the original image, (b, d) attention from the DA prompt (baseline), and (c, e) attention from our DtD strategy. We visualize attention for the final diagnosis tokens (b, c) and for the entire response (d, e). The DtD prompt yields attention maps that are more accurately concentrated on the lesion area, both during the final decision (c) and across the full response (e), illustrating the change in attention allocation.}
    \vspace{-3mm}
    \label{fig:heatmaps_for_different_prompt}
\end{figure*}

Specifically, let the generated sequence have $S$ tokens. At step $t$, after averaging head layers, the decoder gives an attention row $\bar{\mathbf a}^{(t)}\in\mathbb{R}^{T_t}$. Let the image occupy a contiguous token block $\mathcal I=\{s_{\mathrm{img}},\dots,s_{\mathrm{img}}+L_{\mathrm{img}}-1\}$. Define the image-restricted vector, the slice of a token’s attention only over the image tokens, with everything else (text tokens) dropped, as
\begin{equation}
\scalebox{0.88}{$
    \mathbf r^{(t)}\in\mathbb{R}^{L_{\mathrm{img}}},
    \qquad
    \mathbf r^{(t)}[j]=
    \begin{cases}
    \bar a^{(t)}_{\,s_{\mathrm{img}}+j}, & s_{\mathrm{img}}+j\le T_t,\\
    0, & \text{otherwise.}
    \end{cases}
$}
\end{equation}
So $\mathbb{R}^{L_{\mathrm{img}}}$ is just a length-$L_{\mathrm{img}}$ real vector, with one weight per visual token. To check how much the model attends to the final lesion prediction when generating responses (see Fig. \ref{fig:heatmaps_for_different_prompt} (b-c)), the label-token heat vector is $
    \mathbf m_{\text{label}}
    =
    \frac{1}{|\mathcal S_{\text{label}}|}
    \sum_{t\in\mathcal S_{\text{label}}}
    \mathbf r^{(t)}
    \in\mathbb{R}^{L_{\mathrm{img}}}
$
, where $\mathcal S_{\text{label}}$ indexes the tokens that spell the predicted diagnosis. Reshape $\mathbf m_{\text{label}}$ to the token grid and upsample to the image to render the heatmap. For the entire-response attention maps (see Fig. \ref{fig:heatmaps_for_different_prompt} (d-e)), all-tokens heat vector is $
    \mathbf m_{\text{all}}
    =
    \frac{1}{S}
    \sum_{t=1}^{S}
    \mathbf r^{(t)}
    \in\mathbb{R}^{L_{\mathrm{img}}}
$
. By comparing the heatmaps from a baseline prompt (that queries direct answer, Fig.~\ref{fig:heatmaps_for_different_prompt} (b,d)) with those from our DtD prompt (Fig.~\ref{fig:heatmaps_for_different_prompt} (c,e)), we can isolate the effect of our strategy. When the label-token heatmap $\mathbf m_{\text{label}}$ is more concentrated on the lesion, it indicates that the description prompt changes the distribution of attention during the final choice. Similarly, the DtD prompt yields broader yet meaningful attention in the $\mathbf m_{\text{all}}$ heatmap, as the model must scan the image to enumerate pixel-level attributes for the description.

\section{Per-Layer VAR Analysis Across Modalities}
\label{sec:var_all_datasets}

Fig.~\ref{fig:var_layers} presents the per-layer Vision Attention Ratio (VAR) for all six medical imaging datasets under Direct Answer (DA) and Describe-then-Decide (DtD) prompting. The consistent pattern across modalities shows that DtD elevates VAR at every layer, with particularly large gains at the global attention layers (layers 5, 10, 17, 22, 27, 33). Table~\ref{tab:var_summary} provides the numerical summary of mean VAR values and relative improvements.

\section{Attention Entropy Analysis}
\label{sec:entropy_analysis}

To complement the Vision Attention Ratio (VAR) analysis presented in the main text, we examine the entropy of the attention distributions across layers. While VAR measures \emph{how much} attention is directed toward vision tokens, entropy measures \emph{how focused} or \emph{diffuse} that attention is. Lower entropy indicates that the model concentrates its attention on fewer tokens, suggesting more selective visual processing.

Table~\ref{tab:entropy_summary} reports the mean attention entropy under Direct Answer (DA) and Describe-then-Decide (DtD) prompting for six medical imaging datasets. DtD produces lower mean entropy than DA for all six modalities, with reductions ranging from $-0.08$ (PathMNIST, BloodMNIST) to $-0.21$ (OrganAMNIST). The largest decreases are observed for OrganAMNIST ($4.88 \to 4.67$), CheXpert ($4.81 \to 4.61$), and PatchCamelyon ($4.74 \to 4.58$).

\begin{table}[!ht]
\centering
\small
\caption{Mean attention entropy under Direct Answer (DA) vs.\ Describe-then-Decide (DtD).
Lower entropy indicates more focused attention. DtD reduces entropy for all six modalities.}
\label{tab:entropy_summary}
\begin{tabular}{l l c c c}
\toprule
\textbf{Dataset} & \textbf{Modality} & \textbf{DA} & \textbf{DtD}
  & \textbf{$\Delta$} \\
\midrule
Fitzpatrick17k & Dermatology & 4.81 & 4.67 & $-$0.14 \\
CheXpert       & Radiology   & 4.81 & 4.61 & $-$0.20 \\
PatchCamelyon  & Pathology   & 4.74 & 4.58 & $-$0.16 \\
PathMNIST      & Pathology   & 4.77 & 4.69 & $-$0.08 \\
BloodMNIST     & Hematology  & 4.65 & 4.57 & $-$0.08 \\
OrganAMNIST    & CT          & 4.88 & 4.67 & $-$0.21 \\
\bottomrule
\end{tabular}
\end{table}

Fig.~\ref{fig:entropy_heatmap} visualises the per-layer entropy for all six modalities under both prompting strategies. The heatmap confirms that the entropy reduction from DtD is not confined to a few layers but is distributed broadly, with the most pronounced decreases occurring in the global-attention layers that also exhibit the largest VAR increases (see Fig.~\ref{fig:var_layers}).

Taken together with the VAR results, these findings show that DtD not only increases the proportion of attention directed to vision tokens (higher VAR) but also makes that attention more concentrated and selective (lower entropy). This dual effect characterizes how the description prompt changes attention allocation during generation.

\begin{table}[!ht]
\centering
\small
\caption{Mean Vision Attention Ratio (\%) under Direct Answer (DA) vs.\
Describe-then-Decide (DtD). DtD increases the fraction of attention allocated to visual tokens
during generation across the evaluated modalities.}
\label{tab:var_summary}
\begin{tabular}{l l c c c}
\toprule
\textbf{Dataset} & \textbf{Modality} & \textbf{DA} & \textbf{DtD}
  & \textbf{$\Delta$ (Rel.)} \\
\midrule
Fitz17k  & Dermatology & 2.85 & 3.98 & +39.8\% \\
CXR      & Radiology   & 3.02 & 4.24 & +40.2\% \\
PCam     & Pathology   & 2.98 & 3.95 & +32.5\% \\
PathM    & Pathology   & 2.32 & 3.02 & +30.1\% \\
BloodM   & Hematology  & 2.86 & 3.80 & +33.1\% \\
OrganA   & CT          & 2.67 & 3.60 & +34.7\% \\
\bottomrule
\end{tabular}
\end{table}

\begin{figure}[!ht]
\centering
\includegraphics[width=\linewidth]{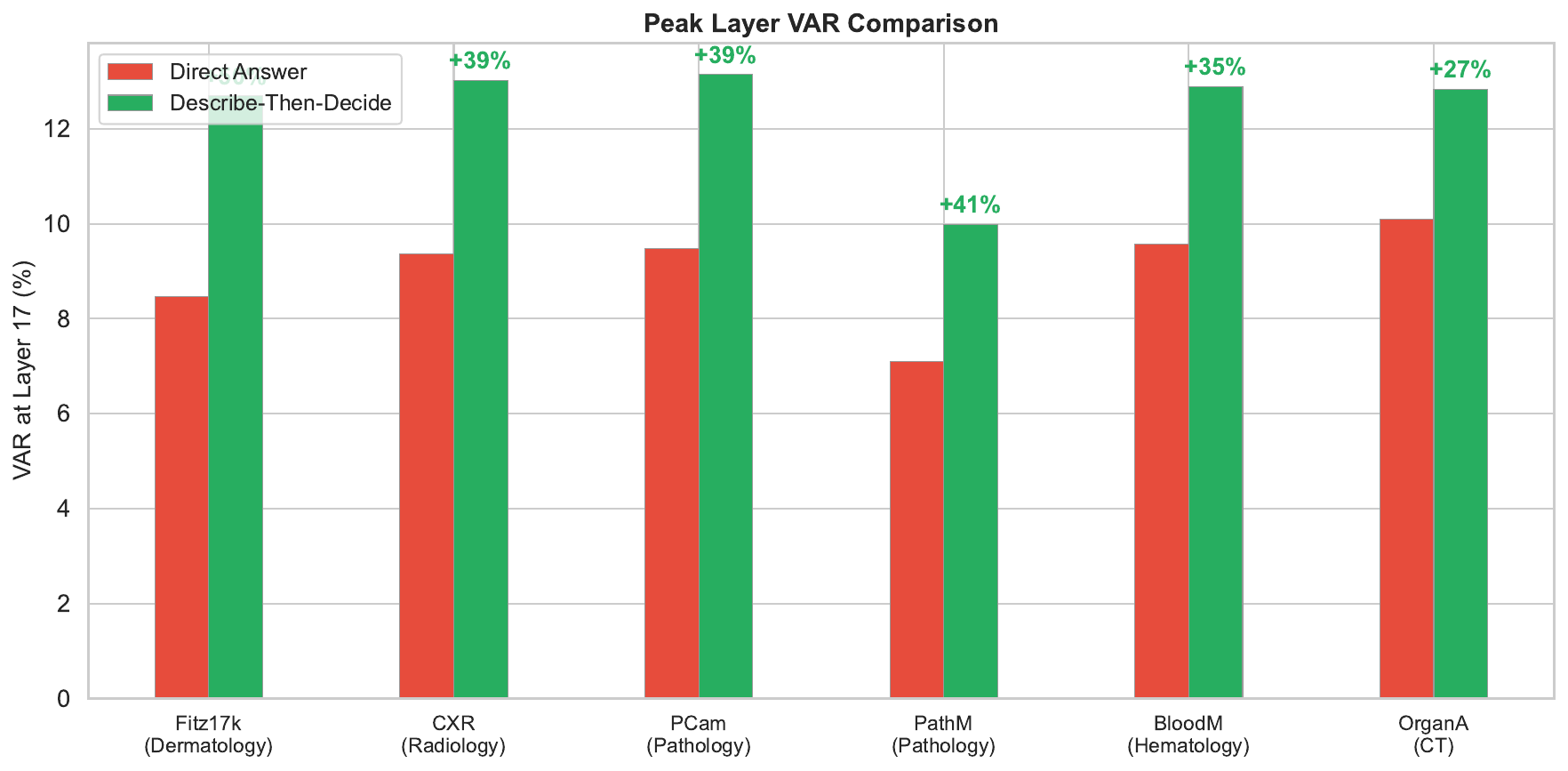}
\caption{Peak-layer Vision Attention Ratio across datasets. DtD consistently achieves higher peak VAR values compared to DA, with the largest improvements observed in radiology and dermatology.}
\label{fig:peak_layer_var}
\end{figure}

\begin{figure*}[!ht]
    \centering
    \includegraphics[width=\linewidth]{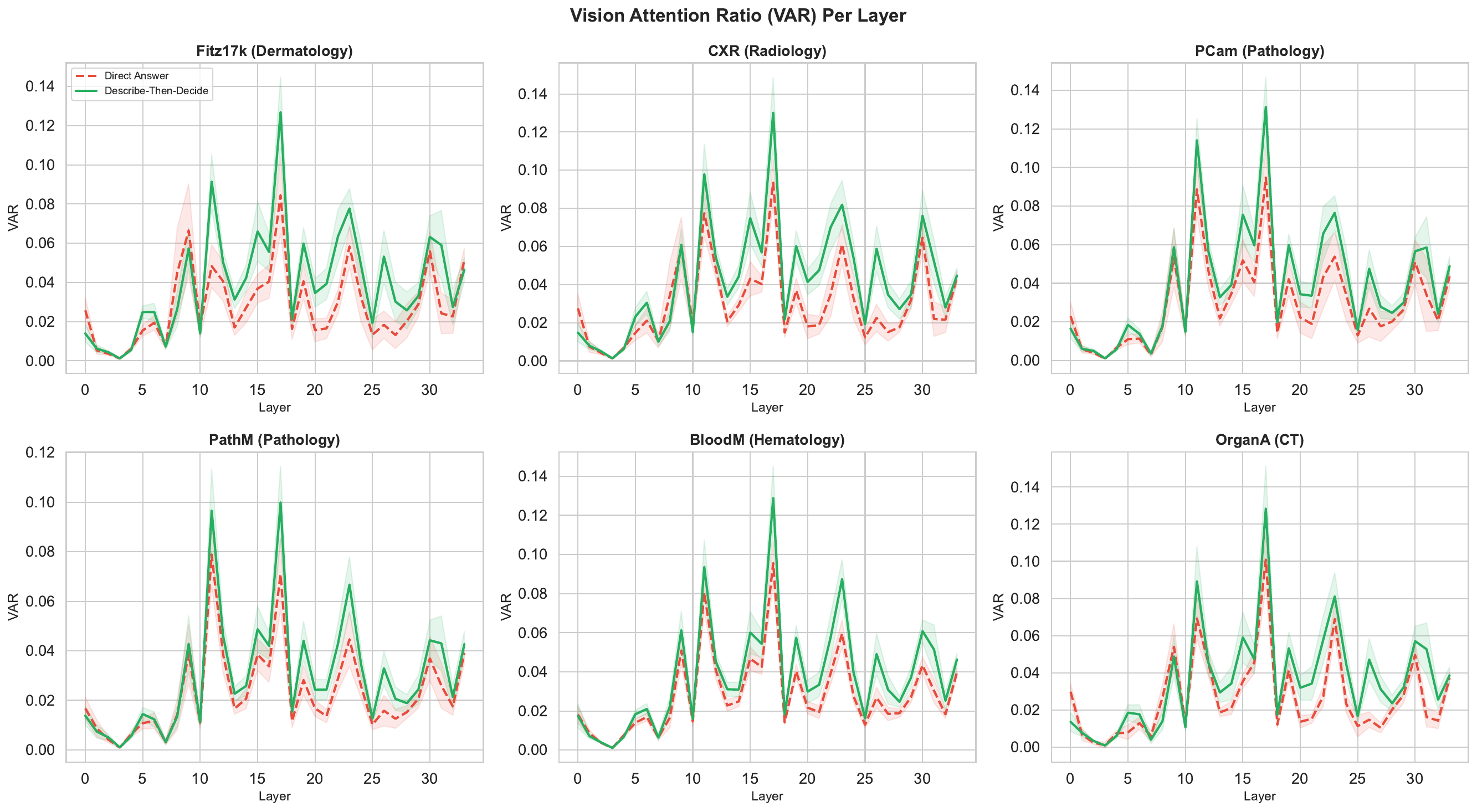}
    \caption{Vision Attention Ratio (VAR) per layer for six medical imaging datasets under Direct Answer (DA, red dashed) vs.\ Describe-then-Decide (DtD, green solid). DtD consistently increases vision attention across all layers and modalities, with the most pronounced gains at global attention layers (5, 10, 17, 22, 27, 33).}
    \label{fig:var_layers}
\end{figure*}

Fig.~\ref{fig:peak_layer_var} further examines the peak-layer VAR values for each dataset, highlighting which layers exhibit the strongest vision attention under each prompting strategy.

\section{Supplementary Experimental Details}
\label{sec:supp_experimental}

\subsection{Abstention Test on Masked Images}

\begin{figure*}[!ht]
    \centering
    \includegraphics[width=1\linewidth]{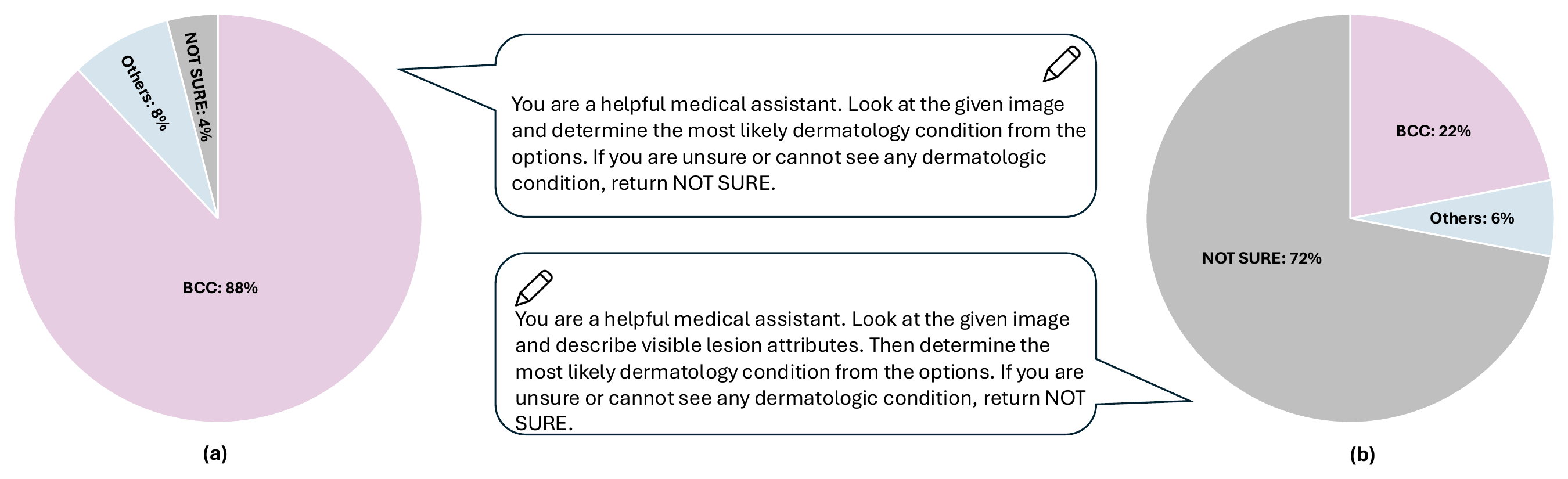}
    \caption{Abstention test on 200 masked images. (a)~DA: the model diagnoses despite masked lesions. (b)~DtD: the model abstains more frequently.}
    \label{fig:masked_image_prediction_distribution}
\end{figure*}

\begin{figure}[!ht]
\centering
\includegraphics[width=\linewidth]{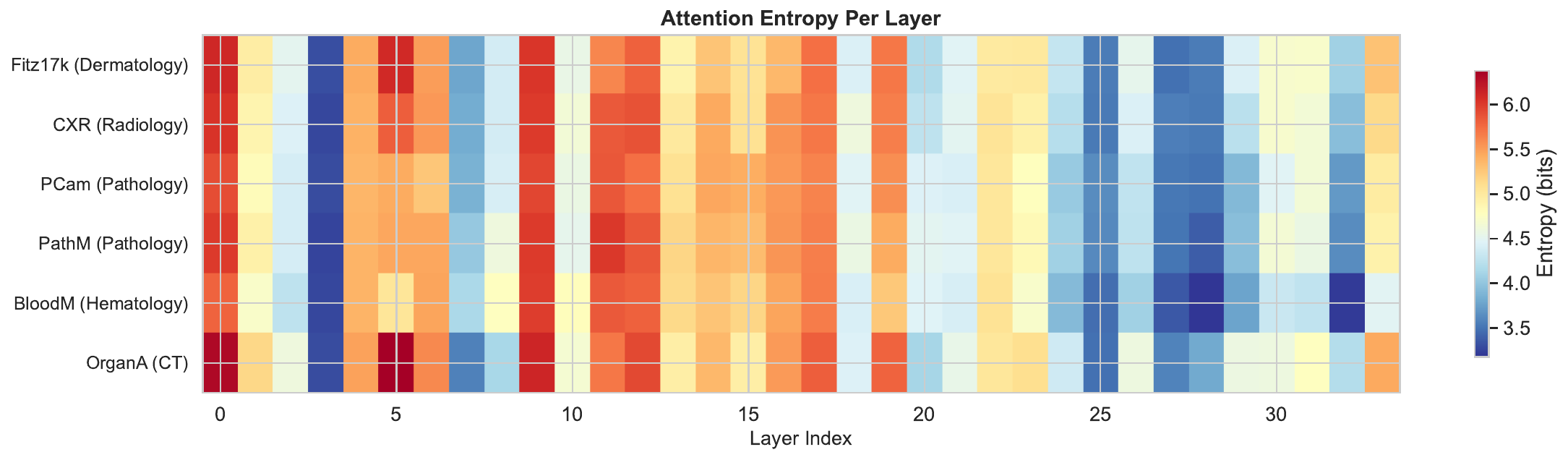}
\caption{Attention entropy heatmap across layers for six medical imaging modalities under Direct Answer (DA) and Describe-then-Decide (DtD) prompting. DtD consistently produces lower entropy (more focused attention) across all modalities and layers.}
\label{fig:entropy_heatmap}
\end{figure}

To quantify the model's reliance on language priors, we masked the primary lesion in 200 test images and evaluated whether MedGemma-4B would abstain from diagnosis. Fig.~\ref{fig:masked_image_prediction_distribution} shows the results under both Direct Answer and Describe-then-Decide prompting. Under DA, the model fails to abstain in 88\% of cases, defaulting to a BCC diagnosis. Under DtD, the abstention rate increases to 72\%, showing that the description prompt increases abstention when the lesion is masked.

\subsection{Description Quality Example}

\begin{figure*}[!ht]
    \centering
    \includegraphics[width=1\linewidth]{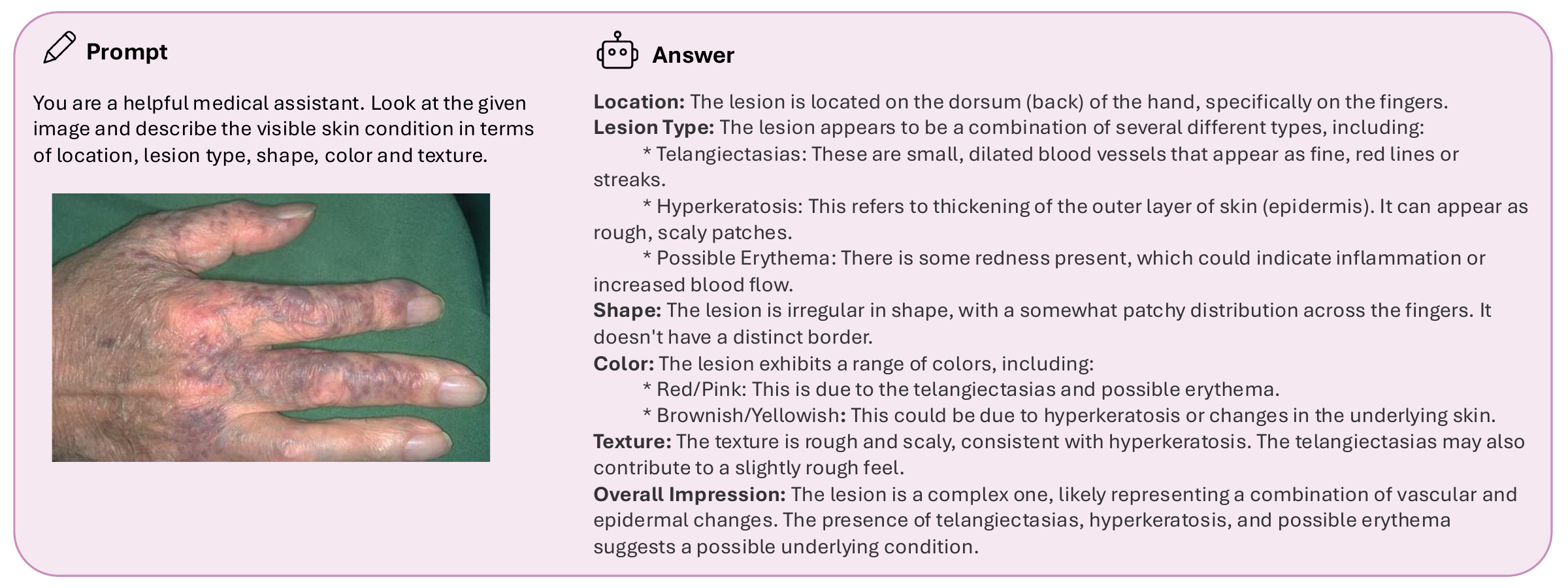}
    \caption{Example description generated by MedGemma-4B. The model produces a detailed, clinically relevant analysis when prompted to describe visible features before diagnosing.}
    \label{fig:description_quality}
\end{figure*}

Fig.~\ref{fig:description_quality} shows an example of a clinical description generated by MedGemma-4B when prompted to describe lesion attributes before diagnosis. The description is detailed and clinically accurate, contrasting sharply with the hallucinated reasoning observed during standard zero-shot classification (cf.\ Fig.~3 in the main text).

\subsection{Zero-Shot Prompt Templates}

Table~\ref{tab:example_prompts} provides the full prompt templates used for each of the three zero-shot prompting strategies evaluated in the main text.

\begin{table}[!ht]
\centering
\caption{Zero-shot prompts used for evaluation.}
\label{tab:example_prompts}
\tiny
\renewcommand{\arraystretch}{1.1}
\begin{tabularx}{\linewidth}{@{}l X@{}}
\toprule
\textbf{Type} & \textbf{Prompt} \\
\midrule
direct answer (baseline) &
You are a helpful medical assistant. Look at the given image and determine the most likely dermatology condition from the options:
\newline (A) Granuloma annulare
\newline (B) Lupus erythematosus
\newline (C) Vitiligo.
\newline ... ...\\
\midrule
in-context &
You are a helpful medical assistant. Look at the given image and determine the most likely dermatology condition from the options:
\newline (A) Granuloma annulare, characterized by smooth skin-colored or pink papules arranged in an annular ring with firm ridge.
\newline (B) Lupus erythematosus, characterized by scaly plaques with follicular plugging and central scarring on sun-exposed skin; acute malar rash appears as flat red butterfly across cheeks sparing folds.
\newline (C) Vitiligo, characterized by sharply bordered milky white macules on face, hands, or genital skin with normal texture and pigment loss.
\newline ... ...\\
\midrule
DtD &
You are a helpful medical assistant. Look at the given image and describe visible lesion attributes (location, lesion type, shape/border, color, texture). Then determine the most likely diagnosis from the options:
\newline (A) Granuloma annulare
\newline (B) Lupus erythematosus
\newline (C) Vitiligo.
\newline ... ...\\
\bottomrule
\end{tabularx}
\end{table}

\section{Implementation Details}
\label{implementation_details}
\subsection{LoRA Fine-tuning}
To ensure a rigorous evaluation of the model's adaptability, we fine-tuned MedGemma-4B-it using Low-Rank Adaptation (LoRA). Given the high memory requirements of 4B parameter models, we employed QLoRA \citep{dettmers2023qlora}, loading the base model in 4-bit precision with bfloat16 compute dtype. We applied LoRA adapters to all linear layers to maximize model expressivity while keeping the number of trainable parameters low. We also explicitly set the embed\_tokens module to be trainable. The optimization was performed using the fused AdamW optimizer with a linear learning rate scheduler and a warmup ratio of 0.03. To determine the optimal configuration, we conducted a grid search over the LoRA rank ($r$), scaling factor ($\alpha$), and learning rate. The search space and selected hyperparameters are detailed in Table~\ref{tab:lora_params}. We trained for 10 epochs with a batch size of 4 and gradient accumulation steps of 4. The final selected configuration (Rank 8, Alpha 16, LR $1e^{-4}$) was chosen based on validation set performance and used for all reported fine-tuning results. Details can be found in Table. \ref{tab:lora_params}.

\begin{table}[!ht]
    \centering
    \caption{Hyperparameter search space and selected configuration for LoRA fine-tuning.}
    \label{tab:lora_params}
    \begin{tabular}{lcc}
        \toprule
        \textbf{Hyperparameter} & \textbf{Search Space} & \textbf{Selected Value} \\
        \midrule
        LoRA Rank ($r$) & $\{4, 8, 16\}$ & $8$ \\
        LoRA Alpha ($\alpha$) & $\{8, 16, 32\}$ & $16$ \\
        Learning Rate & $\{5e^{-5}, 1e^{-4}, 2e^{-4}\}$ & $1e^{-4}$ \\
        LoRA Dropout & - & $0.05$ \\
        Target Modules & - & All Linear Layers \\
        Precision & - & 4-bit (QLoRA) \\
        Optimizer & - & AdamW (Fused) \\
        Training Epochs & - & 10 \\
        \bottomrule
    \end{tabular}
\end{table}

\begin{table}[!ht]
    \centering
    \caption{Full grid search results for LoRA fine-tuning on the Fitzpatrick17k dataset. The optimal configuration selected for the final model is highlighted in bold.}
    \label{tab:lora_grid_search_full}
    \begin{tabular}{cccc}
        \toprule
        \textbf{Rank ($r$)} & \textbf{Alpha ($\alpha$)} & \textbf{Learning Rate} & \textbf{Accuracy (\%)} \\
        \midrule
        4 & 8 & $5\times10^{-5}$ & 42.15 \\
        4 & 8 & $3\times10^{-4}$ & 46.12 \\
        4 & 8 & $4\times10^{-4}$ & 45.88 \\
        \addlinespace
        4 & 16 & $5\times10^{-5}$ & 41.80 \\
        4 & 16 & $3\times10^{-4}$ & 45.05 \\
        4 & 16 & $4\times10^{-4}$ & 44.20 \\
        \addlinespace
        4 & 32 & $5\times10^{-5}$ & 35.60 \\
        4 & 32 & $3\times10^{-4}$ & 39.40 \\
        4 & 32 & $4\times10^{-4}$ & 37.10 \\
        \midrule
        8 & 8 & $5\times10^{-5}$ & 43.50 \\
        8 & 8 & $3\times10^{-4}$ & 46.45 \\
        8 & 8 & $4\times10^{-4}$ & 46.10 \\
        \addlinespace
        8 & 16 & $5\times10^{-5}$ & 44.10 \\
        8 & 16 & $3\times10^{-4}$ & \textbf{47.32} \\ 
        8 & 16 & $4\times10^{-4}$ & 46.95 \\
        \addlinespace
        8 & 32 & $5\times10^{-5}$ & 38.90 \\
        8 & 32 & $3\times10^{-4}$ & 41.25 \\
        8 & 32 & $4\times10^{-4}$ & 40.50 \\
        \midrule
        16 & 8 & $5\times10^{-5}$ & 43.20 \\
        16 & 8 & $3\times10^{-4}$ & 45.90 \\
        16 & 8 & $4\times10^{-4}$ & 45.50 \\
        \addlinespace
        16 & 16 & $5\times10^{-5}$ & 44.80 \\
        16 & 16 & $3\times10^{-4}$ & 46.98 \\
        16 & 16 & $4\times10^{-4}$ & 46.40 \\
        \addlinespace
        16 & 32 & $5\times10^{-5}$ & 43.10 \\
        16 & 32 & $3\times10^{-4}$ & 46.15 \\
        16 & 32 & $4\times10^{-4}$ & 45.20 \\
        \bottomrule
    \end{tabular}
\end{table}

\subsection{Linear Probe}
For feature extraction, we utilized the pre-trained medgemma-4b-it vision encoder in bfloat16 precision. We extracted the global image representation by computing the mean of the last hidden state tokens rather than using a single [CLS] token. The head was trained using the AdamW optimizer and CrossEntropyLoss for 100 epochs with a batch size of 128, as shown in Table. \ref{tab:lp_params}.

\begin{table}[!ht]
    \centering
    \caption{Implementation details and hyperparameters for Linear Probing experiments.}
    \label{tab:lp_params}
    \begin{tabular}{lc}
        \toprule
        \textbf{Hyperparameter} & \textbf{Value} \\
        \midrule
        Vision Encoder & MedSigLIP (Frozen) \\
        Feature Pooling & Global Average Pooling \\
        Precision (Backbone) & \texttt{bfloat16} \\
        Precision (Head) & \texttt{float32} \\
        Optimizer & AdamW \\
        Batch Size & 128 \\
        Epochs & 100 \\
        Learning Rate & $\{3\times10^{-5}, 3\times10^{-4}, 3\times10^{-3}\}$ \\
        \bottomrule
    \end{tabular}
\end{table}

\subsection{Vision Attention Ratio Computation}
\label{sec:var_implementation}
We extract full attention weights using eager attention (disabling Flash Attention) across all 34 transformer layers of MedGemma-4B. For each dataset, we randomly sample 100 test images and compute VAR under both Direct Answer (DA) and Describe-then-Decide (DtD) prompting. MedGemma-4B uses a MedSigLIP vision encoder that produces 256 vision tokens (feature dimension 1152), which are projected into the language model's embedding space. The model uses 8 attention heads per layer. Attention weights are averaged across all heads and all generation steps per layer to produce the final per-layer VAR. All computations are performed in bfloat16 precision on a single NVIDIA A100 GPU.

\subsection{Ablation Study on the Number of Shots}
\label{ablation_shots}
We selected the 8-shot setting based on the analysis presented in Table \ref{table:top5-to-top1}, where 8-shot maximizes the utility of our pipeline. At 8-shot, the vision encoder is strong enough to capture the correct diagnosis in the Top-5 (high recall), but not yet perfect at Top-1 (low precision), creating the ideal scenario for the VLM to apply its reasoning for re-ranking. Beyond 8 shots, the vision encoder becomes self-sufficient, and the added value of VLM reasoning diminishes.

\begin{table}[!ht]
\centering
\small
\caption{Effectiveness of the ``Top-5 to Top-1'' reranking strategy. This table compares three methods: the baseline zero-shot VLM, a standard k-shot linear probe (LP Top-1), and our two-stage approach. In our method, a k-shot linear probe selects the Top-5 candidates, which are then used by the VLM to produce a final prediction. The results show that our two-stage strategy consistently outperforms the direct linear probe's Top-1 accuracy in the low-data regime.}
\label{table:top5-to-top1}
\resizebox{\textwidth}{!}{%
\begin{tabular}{l|c|ccc|ccc|ccc|ccc|ccc}
\toprule
\multirow{2}{*}{\textbf{Dataset}} &
\multirow{2}{*}{\textbf{0-shot Top-1}} &
\multicolumn{3}{c|}{\textbf{1-shot}} &
\multicolumn{3}{c|}{\textbf{2-shot}} &
\multicolumn{3}{c|}{\textbf{4-shot}} &
\multicolumn{3}{c|}{\textbf{8-shot}} &
\multicolumn{3}{c}{\textbf{16-shot}} \\
\cmidrule(lr){3-5}\cmidrule(lr){6-8}
\cmidrule(lr){9-11}\cmidrule(lr){12-14}\cmidrule(lr){15-17}
& &
\textbf{Top1} & \textbf{Top5} & \textbf{0-Shot} &
\textbf{Top1} & \textbf{Top5} & \textbf{0-Shot} &
\textbf{Top1} & \textbf{Top5} & \textbf{0-Shot} &
\textbf{Top1} & \textbf{Top5} & \textbf{0-Shot} &
\textbf{Top1} & \textbf{Top5} & \textbf{0-Shot} \\
\midrule
Derm7pt        & 16.46 & 12.96 & 52.25 & 20.02 & 18.53 & 58.03 & 25.73 & 21.06 & 61.42 & 32.80 & 28.66 & 72.35 & 35.56 & 38.53 & 82.68 & 37.95 \\
eSkinHealth    & 13.35 & 23.51 & 55.75 & 30.98 & 27.99 & 63.84 & 33.67 & 35.04 & 69.70 & 41.22 & 38.80 & 73.96 & 42.05 & 43.04 & 78.41 & 43.17 \\
Fitzpatrick17k & 16.73 & 17.63 & 50.58 & 25.35 & 23.08 & 59.50 & 27.42 & 31.01 & 67.67 & 33.68 & 38.43 & 75.88 & 37.82 & 45.05 & 80.99 & 38.01 \\
SD-260         & 10.81 & 11.48 & 26.89 & 22.31 & 17.90 & 38.61 & 25.02 & 26.97 & 51.92 & 33.50 & 37.12 & 63.61 & 42.33 & 49.01 & 76.72 & 44.72 \\
PAD-UFES-20    & 46.91 & 34.85 & 93.99 & 58.33 & 38.56 & 96.70 & 67.60 & 45.36 & 98.33 & 77.85 & 53.33 & 98.63 & 80.31 & 59.90 & 98.99 & 91.11 \\
\bottomrule
\end{tabular}%
}
\end{table}

\section{Performance Analysis on SkinVL}

\label{skinvl}

To test whether the vision--language gap generalises beyond MedGemma, we repeat the core analysis on SkinVL~\citep{zeng2025mm}, a dermatology VLM built on the LLaVA architecture. Table~\ref{tab:skinvl_gap_analysis} compares zero-shot VLM performance against full fine-tuning and a linear probe on SkinVL's frozen vision encoder across three datasets. The pattern mirrors our MedGemma findings: the linear probe consistently outperforms both the zero-shot VLM and full fine-tuning on every dataset, with the largest gap on Fitzpatrick17k (58.88\% LP vs.\ 10.05\% zero-shot). This confirms that the under-utilisation of visual features is not specific to MedGemma but reflects a broader architectural issue in current medical VLMs.

The cumulative candidate-filtering, clinical-description, and DtD results for SkinVL are provided in Table~\ref{tab:skinvl_main}.

\begin{table}[!ht]
    \centering
    \caption{Performance Gap Analysis on SkinVL. Comparison of the zero-shot VLM performance against full Fine-Tuning (FT) and a Linear Probe (LP) on the vision encoder. Consistent with MedGemma, the Linear Probe significantly outperforms the zero-shot VLM, confirming the under-utilization of visual features.}
    \label{tab:skinvl_gap_analysis}
    \resizebox{\linewidth}{!}{
    \begin{tabular}{l | cc | cc | cc}
        \toprule
         & \multicolumn{2}{c|}{\textbf{VLM Zero-Shot}} & \multicolumn{2}{c|}{\textbf{Fine-Tuning (FT)}} & \multicolumn{2}{c}{\textbf{Linear Probe (LP)}} \\
        \cmidrule(lr){2-3} \cmidrule(lr){4-5} \cmidrule(lr){6-7}
        \textbf{Dataset} & \textbf{ACC} & \textbf{F1} & \textbf{ACC} & \textbf{F1} & \textbf{ACC} & \textbf{F1} \\
        \midrule
        Derm7pt & 12.77 & 15.05 & 44.81 & 22.98 & \textbf{50.16} & \textbf{26.59} \\
        eSkinHealth & 13.76 & 10.33 & 57.74 & 39.24 & \textbf{62.77} & \textbf{42.90} \\
        Fitzpatrick17k & 10.05 & 8.27 & 42.84 & 30.49 & \textbf{58.88} & \textbf{46.22} \\
        \bottomrule
    \end{tabular}
    }
\end{table}
\section{Limitations}
Our main analysis covers five dermatology datasets, with additional backbone comparisons on Derm7pt and Fitzpatrick17k. The experiments in radiology, pathology, hematology, and CT extend the representation and attention analyses; evaluation of the complete pipeline across these specialties remains a next step. The prompt-only interventions use no target labels, whereas Top-5-to-Top-1 depends on a labeled support set and a fitted linear probe. Description quality is measured through model self-evaluation, and independently verified description errors are an important target for further analysis. Attention allocation characterizes model behavior, while causal visual reliance requires additional controlled interventions. Furthermore, exploring how these principles could inform the development of new VLM architectures remains a promising direction.

A comprehensive clinical utility study with multiple experts and inter-rater reliability analysis is a necessary next step for evaluating real-world deployment.

\begin{table}[!ht]
\centering
\caption{Skin Condition Distribution for Fitzpatrick17k}
\small
\begin{tabular}{@{}lrrr@{}}
\toprule
\textbf{Skin Condition} & \textbf{Real Training} & \textbf{Real Test} & \textbf{Synthetic} \\ 
\midrule
Acne (ACN)                  & 92                     & 91                 & 93                 \\
Actinic Keratosis (AK)     & 88                     & 87                 & 164                \\
Allergic Contact Dermatitis (ACD) & 215               & 215                & 181                \\
Basal Cell Carcinoma (BCC)  & 234                    & 234                & 154                \\
Eczema (ECZ)                  & 102                    & 102                & 166                \\
Erythema Multiforme (EM)    & 118                    & 118                & 155                \\
Folliculitis (FOL)           & 171                    & 171                & 114                \\
Granuloma Annulare (GA)      & 106                    & 105                & 148                \\
Keloid (KEL)                 & 78                     & 78                 & 135                \\
Lichen Planus (LP)          & 246                    & 245                & 151                \\
Lupus Erythematosus (LE)    & 205                    & 205                & 172                \\
Melanoma (MEL)               & 130                    & 131                & 155                \\
Mycosis Fungoides (MF)       & 91                     & 91                 & 165                \\
Pityriasis Rosea (PR)       & 96                     & 97                 & 156                \\
Prurigo Nodularis (PN)      & 85                     & 85                 & 152                \\
Psoriasis (PSO)              & 326                    & 327                & 165                \\
Sarcoidosis (SAR)            & 174                    & 175                & 162                \\
Scabies (SCA)                & 170                    & 169                & 176                \\
Squamous Cell Carcinoma (SCC) & 290                    & 291                & 175                \\
Vitiligo (VIT)               & 83                     & 83                 & 161                \\
\midrule
Total                   & 3100                   & 3100               & 3100               \\
\bottomrule
\end{tabular}
\label{tab:fitz_distribution}
\end{table}

\scriptsize
\begin{longtable}{@{} l r r r >{\raggedright\arraybackslash}p{0.60\textwidth} @{}}
\caption{\textbf{Dataset details.}}
\label{dataset_details} \\
\toprule
\textbf{Dataset} & \textbf{\# Cls} & \textbf{Train} & \textbf{Test} & \textbf{Class Names} \\
\midrule
\endfirsthead
\toprule
\textbf{Dataset} & \textbf{\# Cls} & \textbf{Train} & \textbf{Test} & \textbf{Class Names} \\
\midrule
\endhead
\bottomrule
\endfoot
Derm7pt & 14 & 413 & 395 & clark nevus, melanoma, reed or spitz nevus, seborrheic keratosis, basal cell carcinoma, vascular lesion, lentigo, blue nevus, dermal nevus, dermatofibroma, combined nevus, congenital nevus, miscellaneous, recurrent nevus \\
\midrule
eSkinHealth & 24 & 2,714 & 2,676 & buruli ulcer, scabies, yaws, prurigo nodularis, tinea corporis, erysipelas, tinea capitis, leprosy, impetigo, necrotizing fasciitis, contact dermatitis, lichen planus, tinea versicolor, folliculitis, chickenpox, acne, vitiligo, abscess, keratosis, herpes zoster, atopic dermatitis, eczema, lipome, mycetoma \\
\midrule
Fitzpatrick17k & 20 & 3,100 & 3,100 & psoriasis, squamous cell carcinoma, lichen planus, basal cell carcinoma, allergic contact dermatitis, lupus erythematosus, sarcoidosis, folliculitis, scabies, melanoma, erythema multiforme, granuloma annulare, eczema, pityriasis rosea, mycosis fungoides, acne, actinic keratosis, prurigo nodularis, vitiligo, keloid \\
\midrule
SD-260 & 260 & 10,362 & 10,238 & abrasion, acne excoriee, acne keloidalis nuchae, acne vulgaris, acrokeratosis verruciformis, actinic solar damage, actinex treatment, actinic cheilitis, actinic keratosis, cutis rhomboidalis nuchae, actinic favre-racouchot, pigmentation, actinic solar elastosis, solar purpura, telangiectasia, actinic wrinkles, acute eczema, allergic contact dermatitis, alopecia areata, anagen effluvium, androgenetic alopecia, angiofibroma, angiokeratoma, angioma, angular cheilitis, aphthous ulcer, apocrine hydrocystoma, arsenical keratosis, atopic dermatitis, balanitis xerotica obliterans, basal cell carcinoma, beaus lines, beckers nevus, behcets syndrome, benign keratosis, blue nevus, bowenoid papulosis, bowens disease, cafe au lait macule, callus, candidiasis, cellulitis, chalazion, cherry angioma, clubbing of fingers, combined nevus, compound nevus, congenital nevus, contact dermatitis, crowes sign, cutanea larva migrans, cutaneous horn, cutaneous leishmaniasis, cutaneous t-cell lymphoma, cutis marmorata, darier-white disease, dermatofibroma, dermatomyositis, dermatosis papulosa nigra, desquamation, digital fibroma, dilated pore of winer, discoid lupus erythematosus, disseminated actinic porokeratosis, drug eruption, dry skin eczema, dyshidrosiform eczema, dysplastic nevus, eccrine poroma, eczema, epidermal nevus, epidermoid cyst, epithelioma adenoides cysticum, erythema ab igne, erythema annulare centrifigum, erythema craquele, erythema multiforme, exfoliative erythroderma, factitial dermatitis, favre-racouchot, fibroma, fibroma molle, fixed drug eruption, follicular mucinosis, follicular retention cyst, fordyce spots, frictional lichenoid dermatitis, ganglion, geographic tongue, granulation tissue, granuloma annulare, green nail, guttate psoriasis, hailey-hailey disease, half and half nail, halo nevus, hand foot mouth disease, herpes gestationis, herpes simplex virus, herpes zoster, hidradenitis suppurativa, hirsutism, histiocytosis x, hyperkeratosis palmaris et plantaris, hypertrichosis, ichthyosis, ichthyosis vulgaris, id reaction, impetigo, infantile atopic dermatitis, insect bite, intradermal nevus, inverse psoriasis, ischemia, junction nevus, keloid, keratoacanthoma, keratolysis exfoliativa of wende, keratosis pilaris, kerion, koilonychia, kyrles disease, leiomyoma, lentigo maligna melanoma, lentigo simplex, leprosy, leukemia cutis, leukocytoclastic vasculitis, leukonychia, lichen planus, lichen sclerosis et atrophicus, lichen simplex chronicus, lichen spinulosis, linear epidermal nevus, lipoma, livedo reticularis, lymphangioma circumscriptum, lymphocytic infiltrate of jessner, lymphocytoma cutis, lymphomatoid papulosis, mal perforans, malignant melanoma, median nail dystrophy, melasma, metastatic carcinoma, milia, molluscum contagiosum, morphea, mucha-habermann disease, mucous membrane psoriasis, myxoid cyst, nail cosmesis, nail dystrophy, nail nevus, nail psoriasis, nail ridging, nail trauma, neurodermatitis, neurofibroma, neurotic excoriations, nevus cell nevus, nevus comedonicus, nevus incipiens, nevus sebaceous of jadassohn, nevus spilus, nummular eczema, onychogryphosis, onycholysis, onychomycosis, onychoschizia, paronychia, pearly penile papules, pediculosis pubis, pemphigus foliaceus, perioral dermatitis, photodermatitis, pilomatrixoma, pincer nail syndrome, pitted keratolysis, pityriasis alba, pityriasis rosea, pityriasis rubra pilaris, pityriasis versicolor, pityrosporum folliculitis, poikiloderma atrophicans vasculare, pomade acne, porokeratosis of mibelli, port wine stain, pseudofolliculitis barbae, pseudorhinophyma, psoriasis, pterygium inversum unguis, pustular psoriasis, pyoderma gangrenosum, pyogenic granuloma, racquet nail, radiodermatitis, rhinophyma, rosacea, scabies, scalp psoriasis, scar, scarring alopecia, schambergs disease, sebaceous gland hyperplasia, seborrheic dermatitis, seborrheic keratosis, skin tag, solar elastosis, solar lentigo, spindle cell nevus, squamous cell carcinoma (scc), stasis dermatitis, stasis edema, stasis ulcer, steroid acne, steroid atrophy, steroid striae, steroid use, stomatitis, strawberry hemangioma, striae, subacute cutaneous lupus erythematosus, subungual hematoma, superficial actinic porokeratosis, syringoma, systemic lupus erythematosus, terrys nails, thermal burn, tick bite, tinea capitis, tinea corporis, tinea cruris, tinea faciale, tinea incognito, tinea manus, tinea pedis, tinea versicolor, toe deformity, traction alopecia, trichilemmal cyst, trichoepithelioma, trichofolliculoma, trichostasis spinulosa, trichotillomania, tuberous sclerosis, twenty nail dystrophy, ulcer, urticaria, uvl burn, varicella, verruca vulgaris, viral exanthem, virilization, vitiligo, von recklinghausens disease, wart, wound infection, xerosis, x-linked ichthyosis \\
\midrule
PAD-UFES-20 & 6 & 1,134 & 1,164 & actinic keratosis, basal cell carcinoma, melanoma, nevus, seborrheic keratosis, squamous cell carcinoma \\
\bottomrule
\end{longtable}
\scriptsize
\begin{longtable}{@{} l >{\raggedright\arraybackslash}p{0.78\textwidth} @{}}
\caption{\textbf{Skin condition descriptions used in this study.}}
\label{tab:condition_descriptions} \\

\toprule
\textbf{Condition} & \textbf{Description} \\
\midrule
\endfirsthead

\toprule
\textbf{Condition} & \textbf{Description} \\
\midrule
\endhead

\bottomrule
\endfoot

acne & Face/chest/back comedones and inflamed papules/pustules ±nodules; red or skin-colored with black/white heads; oily, ±crust. \\
actinic keratosis & Sun-exposed rough, scaly flat/slightly raised papule <1 cm; pink/red/brown; gritty sandpaper feel. \\
allergic contact dermatitis & At contact sites, ill-defined pink-red (darker on dark skin) patches/plaques ±vesicles/edema; weepy/crusty/scaly surface. \\
basal cell carcinoma & Sun-exposed pearly/waxy papule or thin scaly patch with rolled edge ±central ulcer; translucent or brown/black; smooth/shiny ±crust. \\
eczema & Flexural dry itchy ill-defined patches/plaques ±tiny vesicles; red/pink or purple/gray on dark skin; flaky ±lichenified. \\
erythema multiforme & Acral target lesions—round 1–3 cm with dark center, pale ring, red outer rim—mostly flat ±blister. \\
folliculitis & Hair-bearing sites with clustered 2–5 mm follicle-centered pustules/red papules; red/darker base with white/yellow pus; dome-shaped ±crust. \\
granuloma annulare & Hands/feet/wrists/ankles smooth firm non-scaly papules forming annular rings; skin-colored/pink/red (purple on dark skin). \\
keloid & Over scars on chest/shoulders/earlobes etc., shiny firm hairless raised irregular growth extending beyond wound; pink/red or darker. \\
lichen planus & Wrists/ankles etc. flat-topped polygonal 2–10 mm violaceous papules/plaques with fine white Wickham striae; shiny ±scale. \\
lupus erythematosus & Malar ‘butterfly’ smooth pink rash ±discoid coin-shaped scaly scarred plaques on scalp/ears; red or hyperpigmented; rough if discoid. \\
melanoma & Anywhere (palms/soles/nails in dark skin) asymmetric lesion with irregular borders, color variegation, often >6 mm; becomes raised/crusted/ulcerated. \\
mycosis fungoides & Non-sun areas with dry scaly patches → thicker scaly plaques (±smooth tumor nodules); pink-red to brown/darker; irregular. \\
pityriasis rosea & Trunk herald patch then multiple smaller ovals along skin lines; pink/salmon (gray/brown/purple on dark skin) with fine collarette scale. \\
prurigo nodularis & Reachable areas with multiple very itchy 1–3 cm firm nodules, often crusted/scabbed on top; pink/red/brown/black; thick/rough with excoriations. \\
psoriasis & Elbows/knees/scalp/lower back well-demarcated plaques with thick silvery/gray scale; pink/red or purple/dark brown; dry/flaky (Auspitz sign). \\
sarcoidosis & Face/shins/scars with smooth firm plaques/nodules/patches; purplish/red-brown or lighter/darker areas; rubbery; shin nodules are tender. \\
scabies & Finger webs/wrists/waist/genitals etc. with 5–15 mm wavy burrows plus clustered 1–2 mm itchy papules/vesicles; excoriated/crusted. \\
squamous cell carcinoma & Sun-exposed or scarred sites with firm scaly/crusted nodule/plaque/ulcer ±raised border/central depression; pink/red or darker; rough, may bleed. \\
vitiligo & Sharply bordered depigmented patches (often symmetric) on face/hands/feet/genitals; chalk-white contrast; normal texture without scale. \\
nevus & Anywhere well-circumscribed round/oval macule/papule/nodule with smooth borders; uniform tan/brown/black or skin-colored; smooth flat or dome-raised. \\
seborrheic keratosis & Anywhere except palms/soles ‘stuck-on’ waxy papule/plaque with sharp borders; tan to dark brown/black or mixed; warty/greasy with keratin plugs. \\
buruli ulcer & Limb lesion evolves from painless nodule/plaque/edema to undermined necrotic ulcer; circular/irregular with yellow-white base and violaceous/brown edge; moist ±slough. \\
yaws & Leg/foot raspberry-like papilloma ±ulcer then multiple papules/plaques/shallow ulcers; dome-lobulated; bright red → red-brown/yellow; verrucous/granular ±crust. \\
tinea capitis & Scalp alopecic scaly ring-like patch with broken hairs/black dots ±boggy kerion or yellow scutula; gray-white/red-brown; fine scale or boggy crust. \\
tinea corporis & Exposed skin annular plaques with raised scaly active border and central clearing; polycyclic/serpiginous; rim scaly, center smoother. \\
erysipelas & Lower legs/face sharply demarcated lobulated cellulitic plaque ±tense bullae; fiery red/violaceous; smooth tense shiny peau d’orange. \\
leprosy & Cool sites hypopigmented/coppery patches with sensory loss or plaques/nodules/diffuse infiltration; round/oval with raised rim; dry hairless/anhidrotic or thick shiny. \\
necrotizing fasciitis & Rapidly enlarging ill-defined limb/trunk plaque with dusky patches, ecchymoses and flaccid bullae → black necrotic eschar; tense shiny then leathery. \\
impetigo & Perioral/exposed erosions with honey-colored crust after vesicles/pustules (bullous form flaccid blisters); round/oval, coalescent; moist → sticky crust. \\
tinea versicolor & Upper trunk/shoulders/neck flat hypo- or hyperpigmented macules/patches with fine powdery ‘bran’ scale accentuated by scraping; coalescing map-like. \\
varicella & Face/trunk then scalp/limbs crops of itchy 2–4 mm thin-walled vesicles on pink/dark base → crusted scabs; discrete ±coalescent; glistening then flaky. \\
abscess & Axillae/buttocks/groin etc. tender dome-shaped red-violaceous nodule that becomes fluctuant with pointing yellow/white center; shiny tense skin; purulent drainage after rupture. \\
atopic dermatitis & Flexures/hands/eyelids etc. itchy ill-defined papules→plaques with acute weeping/crust or chronic lichenification; pink-red or purple-brown/gray; dry rough. \\
keratosis pilaris & Outer arms/thighs/cheeks tiny follicular keratin plugs (‘goose-bump’ papules) in indistinct patches; skin-colored to red; rough sandpapery dry feel. \\
herpes zoster & Single-dermatome band of clustered clear vesicles on pink/purple base not crossing midline → pustules/crusts; 2–4 mm; tense/glossy then adherent crust. \\
lipoma & Trunk/neck/limbs soft smooth mobile subcutaneous dome/oval nodule with normal overlying color; rubbery/doughy consistency. \\
mycetoma & Foot/leg/hand firm lobulated subcutaneous mass with multiple draining sinus openings extruding colored grains; overlying skin normal→hyperpigmented; crusted seepage. \\
lymphatic filariasis & Lower legs/feet (±arms/genitals) chronic lymphedema to elephantiasis with column-like limb, bulbous foot and verrucous corrugated plaques; hyper/hypopigmented; thick ‘mossy’ hyperkeratosis. \\

\end{longtable}

\noindent\begin{minipage}{\linewidth}
\centering
\scriptsize
\captionof{table}{Visual checklists for Basal Cell Carcinoma (BCC) and Lupus Erythematosus (LE), used to probe MedGemma’s conceptual knowledge of misclassified conditions.}
\label{tab:bcc_desc}
\renewcommand{\arraystretch}{1.15}
\begin{tabularx}{\textwidth}{@{} l l X @{}}
\toprule
\textbf{Condition} & \textbf{Feature} & \textbf{Details} \\
\midrule
\multirow{5}{*}{\textbf{BCC}}
& Location & Sun-exposed areas: face (nose, cheeks, forehead), ears, neck, scalp, upper chest/shoulders, back, hands, arms \\
& Morphology & ``Pearly’’/waxy bump; papillary (dome-shaped); nodular (firm); superficial infiltrating (flat, scaly patch); morpheaform (scar-like plaque); micropapillary; infiltrating (flat, scaly, mimics other dermatoses) \\
& Shape & Often irregular with poorly defined borders; typically asymmetrical; evolving in size, shape, or color \\
& Color & Pink or red; may be white/waxy; brown/tan or black; blue/purple hues if ulcerated \\
& Texture & Smooth or waxy; may be scaly or crusted; ulcerated surface can occur \\
\midrule
\multirow{5}{*}{\textbf{LE}}
& Location & Sun-exposed areas: face, ears, neck, chest, upper arms; mucous membranes (oral cavity, nose, eyelids); scalp, nails, genital area \\
& Morphology & Malar (butterfly) rash: flat/raised erythema across cheeks and nasal bridge; discoid: coin-shaped scarring plaques with hypo-/hyperpigmentation; photosensitivity rash; oral ulcers; livedo reticularis; vasculitis (purpura, petechiae); alopecia; Raynaud’s phenomenon \\
& Shape & Malar: butterfly-shaped with central clearing; discoid: coin-shaped plaques; oral ulcers: round/oval; livedo: reticular (net-like); purpura: pinpoint to macular \\
& Color & Malar: red with central clearing; discoid: red, heals with pigment changes; oral ulcers: red to violaceous; livedo: reddish-blue; purpura: red or purple \\
& Texture & Malar: flat or slightly raised; discoid: raised, scaly, may crust or scar; oral ulcers: smooth base; livedo: smooth with reticular pattern \\
\bottomrule
\end{tabularx}
\label{tab:le_desc}
\end{minipage}

\end{document}